\documentclass[letterpaper]{article} % DO NOT CHANGE THIS
\usepackage{aaai2026}  % DO NOT CHANGE THIS
\usepackage{times}  % DO NOT CHANGE THIS
\usepackage{helvet}  % DO NOT CHANGE THIS
\usepackage{courier}  % DO NOT CHANGE THIS
\usepackage[hyphens]{url}  % DO NOT CHANGE THIS
\usepackage{graphicx} % DO NOT CHANGE THIS
\usepackage{framed}
\usepackage{natbib}  % DO NOT CHANGE THIS AND DO NOT ADD ANY OPTIONS TO IT
\usepackage{caption} % DO NOT CHANGE THIS AND DO NOT ADD ANY OPTIONS TO IT
\usepackage{algorithm}
\usepackage{algorithmic}

\usepackage{newfloat}
\usepackage{listings}
\DeclareCaptionStyle{ruled}{labelfont=normalfont,labelsep=colon,strut=off} % DO NOT CHANGE THIS
\floatstyle{ruled}
\newfloat{listing}{tb}{lst}{}
\floatname{listing}{Listing}
\title{Detecting Unobserved Confounders: A Kernelized Regression Approach}
\author{
    Yikai Chen\textsuperscript{\rm 1}\equalcontrib, 
    Yunxin Mao\textsuperscript{\rm 1}\equalcontrib, 
    Chunyuan Zheng\textsuperscript{\rm 2},
    Hao Zou\textsuperscript{\rm 3}, 
    Shanzhi Gu\textsuperscript{\rm 1}, \\ 
    Shixuan Liu\textsuperscript{\rm 1}, 
    Yang Shi\textsuperscript{\rm 4}, 
    Wenjing Yang\textsuperscript{\rm 1},
    Kun Kuang\textsuperscript{\rm 5},
    Haotian Wang\textsuperscript{\rm 1}\thanks{Corresponding Author.}
}
\affiliations{
    \textsuperscript{\rm 1}College of Computer Science, National University of Defense Technology, Changsha, China\\
    \textsuperscript{\rm 2}School of Mathematical Sciences, Peking University, Beijing, China\\
    \textsuperscript{\rm 3}Department of Computer Science, Tsinghua University, Beijing, China\\
    \textsuperscript{\rm 4}School of Computer Science, Peking University, Beijing, China\\
    \textsuperscript{\rm 5}College of Computer Science and Technology, Zhejiang University, Hangzhou, China
    
    chenyikai1998@nudt.edu.cn
}

\usepackage{amsmath}
\usepackage{amssymb}
\usepackage{dsfont}
\usepackage{subcaption}
\usepackage{amsthm}
\usepackage{booktabs}
\usepackage{multirow}

\newtheorem{theorem}{Theorem}

\newtheorem{remark}{Remark}
\newtheorem*{Proof}{Proof}
\newtheorem{assumption}{Assumption}

\begin{document}

\maketitle

\begin{abstract}
Detecting unobserved confounders is crucial for reliable causal inference in observational studies. Existing methods require either linearity assumptions or multiple heterogeneous environments, limiting applicability to nonlinear single-environment settings. To bridge this gap, we propose Kernel Regression Confounder Detection (KRCD), a novel method for detecting unobserved confounding in nonlinear observational data under single-environment conditions. KRCD leverages reproducing kernel Hilbert spaces to model complex dependencies. By comparing standard and higher-order kernel regressions, we derive a test statistic whose significant deviation from zero indicates unobserved confounding. Theoretically, we prove two key results: First, in infinite samples, regression coefficients coincide if and only if no unobserved confounders exist. Second, finite-sample differences converge to zero-mean Gaussian distributions with tractable variance. Extensive experiments on synthetic benchmarks and the Twins dataset demonstrate that KRCD not only outperforms existing baselines but also achieves superior computational efficiency.
\end{abstract}


\section{Introduction}
Causal effect estimation plays a central role in diverse high-stakes domains, such as healthcare~\citep{prosperi2020causal, dahabreh2024causal, wang2024graphcl}, public policy~\citep{matthay2022causal}, and financial decision-making~\citep{kuhne2022causal, vanderschueren2024operational}. While randomized controlled trials (RCTs) remain the gold standard for causal inference, their practical deployment is often hindered by ethical concerns or limited scalability. Consequently, observational studies have emerged as a primary alternative, valued for their broader external validity and reduced cost \citep{pearl2009causality,listl2016causal}.

However, the reliability of observational studies critically depends on the \textit{unconfoundedness assumption}---the requirement that all variables confounding the treatment-outcome relationship are fully measured and accounted for. When the unconfoundedness assumption does not hold, spurious correlations will undermine causal conclusions drawn from observational data~\citep{angrist1996identification}. Intuitively, before mitigating the effect of unobserved confounders~\citep{lee2010regression}, we need first to determine whether such confounders exist, which is the main focus of this paper. Thus, detecting the presence of unobserved confounders plays a foundational role in guiding subsequent methodological decisions.

Existing methods for detecting unobserved confounding can be broadly divided into two categories. The first one requires linear model assumptions, which imposes linear functional relationships among all variables~\citep{schultheiss2024higher, janzing2018detecting, liu2018confounder}. While straightforward, these assumptions are not testable and severely hinder detection performance on nonlinear data. The second approach requires multiple environments with mechanism-varying Structural Causal Models (SCMs), leveraging causal mechanism shifts to detect confounders via propagation signatures~\citep{karlsson2023detecting, mameche2024identifying, reddy2024detecting}. However, in practice, it is challenging to obtain sufficiently heterogeneous environments satisfying the required causal invariance and mechanism independence assumptions. 
Note that both these two types of methods make assumptions on the data structure, which raises an important open question:
\begin{framed}
    \textit{Can unobserved confounders be detected in non-linear settings within a single data environment?}
\end{framed}

To this end, we propose an effective method named ``{\bf K}ernel {\bf R}egression {\bf C}onfounder {\bf D}etection''~(KRCD), which detects unobserved confounders on the nonlinear observational data drawn from a single environment. Specifically, the proposed KRCD method consists of three main steps: nonlinear structure modeling, kernelized regression, and hypothesis construction. To characterize the non-linear data structure, our KRCD first maps the treatment variable~($T$), the outcome variable~($Y$) and observed covariates~($X$) to reproducing kernel Hilbert space (RKHS), enabling a linear inner-product representation of the underlying nonlinear structural functions. Subsequently, the proposed KRCD regresses $Y$ on $(T,X)$ in the RKHS twice: the first is a standard kernelized regression, while the second is a higher-order regression weighted by $\|T,X\|^2$, designed to capture nonlinear dependencies indicative of unobserved confounding. Finally, we construct the hypothesis test, which rejects the presence of unobserved confounders if the normalized difference between the two regression coefficients becomes sufficiently small. The main contributions of this paper can be summarized as follows:

\begin{itemize}
\item We investigate the problem of detecting unobserved confounders from nonlinear data in a single-environment setting. To address this, we propose KRCD, a novel method that compares standard kernel regression and higher-order kernel regression in RKHS, enabling efficient detection of unobserved confounding.
\item Theoretically, we show that standard kernel regression and higher-order kernel regression coefficients coincide if and only if no unobserved confounders exist in infinite samples. For finite samples, their difference asymptotically converges to a zero-mean Gaussian distribution with tractable asymptotic variance.
\item Through extensive experiments on synthetic benchmark datasets and the Twins dataset, our method stably outperforms the baselines in detecting unobserved confounders. 
\end{itemize}

\section{Related Work}
\subsection{Detection of Unobserved Confounders in Single-Environment Settings}
Traditional approaches to detect unobserved confounding---including instrumental variable (IV) regression \citep{angrist1996identification} and the Durbin-Wu-Hausman test \citep{hausman1978specification}---rely on external instruments or estimator comparisons \citep{wang2022estimating, kou2025label}. Subsequent innovations like ROCD \citep{schultheiss2024higher} derived from reweighting strategies \citep{buja2019models_a,buja2019models_b} bypass instrument requirements by leveraging non-Gaussian higher-order moments, offering a resampling-free OLS/HOLS comparison framework~\citep{wang2023treatment}. For high-dimensional linear models, spectral methods such as \citep{liu2018confounder, janzing2018detecting} detect confounding through spectral analysis of regression-induced associations or deviations in covariance structures, with \citep{janzing2018detecting} explicitly linking confounding to multivariate linear model perturbations. Beyond linearity, early efforts to address nonlinear confounding relied on strong structural assumptions (e.g., nonparametric IV \citep{newey2003instrumental}) or specific non-Gaussian distributions (e.g., LiNGAM \citep{shimizu2006linear}), yet lacked general nonparametric tests for unobserved confounders. While these methods advance instrument-free detection, they remain fundamentally constrained to linear models---a limitation overcome by our KRCD method, which extends confounder detection to nonlinear relationships.

\subsection{Detection of Unobserved Confounders in Multi-Environment Settings}
Beyond linear constraints, recent methods exploit causal mechanism shifts across environments to detect unobserved confounding without causal sufficiency assumptions. Theoretical work established testable conditional independence violations indicating confounding under shifts \citep{karlsson2023detecting}. Subsequent studies developed operational frameworks: information-theoretic detection measures \citep{mameche2024identifying} and nonparametric quantification of confounding strength via mechanism shifts \citep{reddy2024detecting}. These approaches treat environmental variations as natural experiments to derive confounding signals~\citep{yin2023coco, wang2025simprof, wang2023out}. Joint causal inference \citep{mooij2020joint} and causal discovery under heterogeneity \citep{perry2022causal} also facilitate confounder detection. However, such methods typically require explicit distribution shifts or auxiliary experimental data \citep{kallus2018removing, hu2024sequential}, limiting their use in purely observational, single-context settings. In contrast, KRCD achieves robust confounding detection in single-environment observational data while maintaining nonparametric flexibility, eliminating the need for multi-source data.

\subsection{Kernelized Causal Inference}
The theoretical foundation of Reproducing Kernel Hilbert Spaces (RKHS) was established by Aronszajn's seminal work on the reproducing property \citep{aronszajn1950theory}, with the Representer Theorem \citep{kimeldorf1971some} enabling efficient regularization via kernel expansions. This framework supports three key causal strategies: (1) Kernel mean embeddings (KME) \citep{hazlett2016kernel,muandet2017kernel} for covariate balancing assume no unobserved confounders; (2) RKHS-based IV regression \citep{singh2019kernel,mastouri2021proximal} handles observed confounding but requires external instruments; (3) Latent state identification \citep{zhang2015multi,pfister2018kernel} infers unmeasured variables without explicit confounder detection. Despite enhancing nonparametric flexibility, these methods remain limited by causal sufficiency assumptions or auxiliary data dependencies.

Critically, no existing RKHS approach directly detects nonlinear unobserved confounding in standard observational data. Current techniques either ignore latent biases (KME), depend on untestable instruments (kernel IV), or lack bias quantification (independence tests)~\citep{li2023multiple, wang2025debiased, wang2025gaussian}. To address this gap, we unify RKHS theory with the ROCD framework \citep{schultheiss2024higher} through a novel Representer Theorem variant. This unification enables data-adaptive kernel bases for nonlinear confounding detection, forming our KRCD method. KRCD pioneers direct detection of nonlinear unobserved confounders by embedding ROCD's higher-order moment structures into RKHS---requiring neither instruments nor multi-environment data.

\section{Preliminaries and Problem Setup}

In this paper, we denote the treatment, observed covariates, outcome, and the unobserved confounder as $T$, $X$, $Y$ and $U$, respectively. We follow the potential outcome framework and characterize the potential outcome of $Y$ under the assignment $T=t$ as $Y(t)$. We denote the random variables by uppercase letters~(e.g., $T$ and $Y$) and their realizations by lowercase letters~(e.g., $t$ and $y$). 

\subsection{Previous Problem Setup}
Existing approaches for detecting unobserved confounders bifurcate into two constrained paradigms:
\begin{itemize}
\item \textbf{Linear constraints:} Single-environment methods universally require parametric functional assumptions to achieve identifiability. This precludes their application to general nonlinear systems with complex dependencies.
\item \textbf{Multi-environment dependency:} Multi-environment methods universally require observable causal mechanism shifts in SCMs. This fundamentally precludes application to single-environment observational studies.
\end{itemize}

\subsection{Our Problem: Detecting Unobserved Confounders in Nonlinear Single-Environment~(NSE)}
We aim to \textit{detect the existence of unobserved confounders $U$ in nonlinear single-environment settings}. Under the null hypothesis $H_0$, no unobserved confounders exist, specified by the structural equations: 
\begin{equation}
H_0: \left\{
\begin{aligned}
T &= f_T(X) + \epsilon_T, \\
Y &= g(T, X) + \epsilon_Y. \label{eq1}
\end{aligned}
\right.
\end{equation}
%with nonlinear functions $f_T$, $g$, and exogenous noise terms $\epsilon_T$, $\epsilon_Y$. 
Conversely, under the alternative hypothesis $H_1$, unobserved confounders are present, characterized by:
\begin{equation}
H_1: \left\{
\begin{aligned}
T &= f_T(X) + h_T(U) + \epsilon_T, \\
Y &= g(T, X) + h_Y(U) + \epsilon_Y, \label{eq2}
\end{aligned}
\right.
\end{equation}
where $f_T$, $g$, $h_T$, and $h_Y$ are nonlinear functions, $\epsilon_T$ and $\epsilon_Y$ are exogenous noise terms. For subsequent analysis, we define $Z = (T,X)^\top$ with $\mathbb{E}[\epsilon_T] = \mathbb{E}[\epsilon_Y] = 0$.

\begin{remark}[Distinction from Previous Problem Setups]
Our proposed method differs fundamentally from prior works in two key aspects:

(1) \textbf{Nonparametric formulation}: We do not assume parametric functional forms (e.g., linearity) for $f_T$, $g$, $h_T$, $h_Y$, or distributions of $\epsilon_T$, $\epsilon_Y$, making our approach applicable to general nonlinear systems.

(2) \textbf{Single-environment setting}: Unlike methods relying on multiple environments (e.g., stochastic mechanism shifts, or explicit interventions), we address the more challenging scenario where only \textit{observational data from a single environment} is available.

This positions our work as the first nonparametric detection for unobserved confounders using single-environment observational data.
\end{remark}

\section{Kernel Regression for Confounder Detection}
In this section, we propose {\bf K}ernel {\bf R}egression for {\bf C}onfounder {\bf D}etection (KRCD), a novel and powerful method that leverages Reproducing Kernel Hilbert Space (RKHS) theory to extend the Residual Orthogonality Condition for Diagnostics (ROCD) framework for nonparametric detection of unobserved nonlinear confounding in high-dimensional settings. This approach significantly enhances the diagnostic power of moment-based methods beyond linear relationships.

\begin{figure*}[t]
  \centering
  \includegraphics[width=\textwidth]{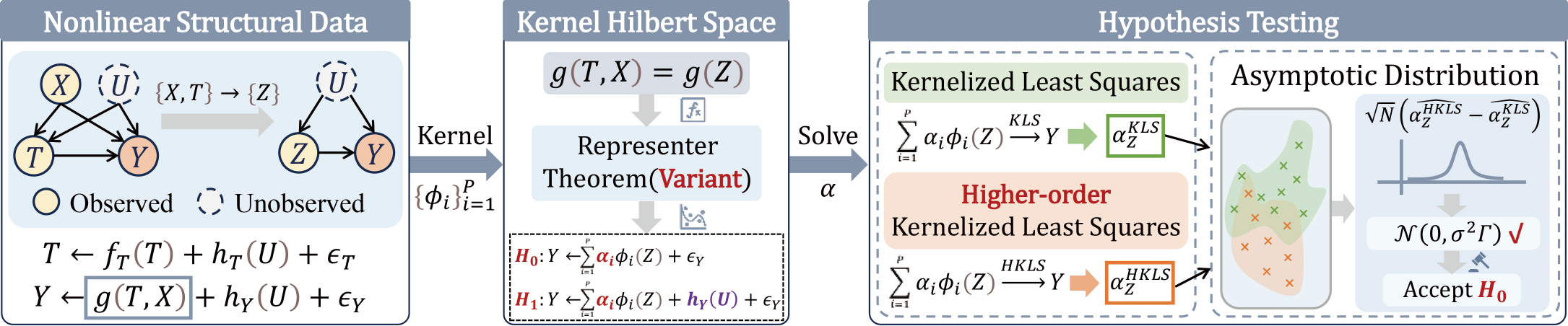} 
  \caption{Workflow for Proposed Method (KRCD).}
  \label{fig1}
\end{figure*}

\subsection{Motivation}
In the NSE setting with potential nonlinear relationships, conventional linear detection methods face fundamental limitations. This motivates our reexamination of the ROCD framework's core premise:

\noindent {\bf Regression-Oriented Confounder Detection (ROCD). } 
ROCD exploits the asymptotic equivalence of OLS and Higher-order least squares (HOLS) estimators under correct linear specification. Systematic differences indicate misspecification (e.g., unobserved confounding). The method:
\begin{itemize}
\item Computes OLS and HOLS coefficients ($Y$ on $X,T$).
\item Tests coefficient equality via standardized differences; significance implies unobserved confounders.
\end{itemize}

However, ROCD is \textit{limited to linear regression}, failing for nonlinear data. To extend this paradigm to nonlinear settings, we define two types of least squares estimators for the nonlinear function $g(Z)$:
\begin{equation}
g^{OLS} = \underset{g \in \mathcal{H}}{\operatorname{argmin}} \, \mathbb{E}\left[ \left\| Y - g(Z) \right\|^2  \right], \label{equ:main:g1}
\end{equation}
\begin{equation}
g^{HOLS} = \underset{g \in \mathcal{H}}{\operatorname{argmin}} \, \mathbb{E}\left[ \left\| Y - g(Z) \right\|^2 \|Z\|^2 \right], \label{equ:main:g2}
\end{equation}
where $\mathcal{H}$ is an appropriate function space.

\noindent {\bf Challenges.} In linear settings where $g$ resides in finite-dimensional Euclidean space, $g^{OLS}$ and $g^{HOLS}$ admit explicit vector representations. However, for nonlinear $g$ in infinite-dimensional function spaces, representing $g$ for comparative analysis and downstream detection remains a fundamental theoretical and computational challenge.

\subsection{Representer for Functional Regression}
To address this challenge, we leverage the Representer Theorem in Reproducing Kernel Hilbert Spaces. This approach transforms the infinite-dimensional optimization problem into a finite-dimensional convex optimization over coefficients, enabling the capture of complex nonlinear dependencies indicative of unobserved confounding.

\noindent {\bf Reproducing Kernel Hilbert Space (RKHS).}~\citep{aronszajn1950theory} A RKHS $\mathcal{H}_k$ is a function space with reproducing kernel $k$, satisfying $\langle f, k(\cdot, x) \rangle_{\mathcal{H}_k} = f(x)$. Its feature map $\phi(x) = k(\cdot, x)$ embeds inputs into $\mathcal{H}_k$.

\noindent {\bf Representer Theorem.}~\citep{kimeldorf1971some} For $g$ in RKHS $\mathcal{H}_k$ induced by kernel $k$, the minimizer of a regularized empirical risk admits the form: $\hat{g}(\cdot) = \sum_{i=1}^N \alpha_i \phi_i(\cdot)$, where $\phi_i(\cdot) = k(x_i, \cdot)$, $\{x_i\}_{i=1}^N$ are training sample points and $\alpha_i \in \mathbb{R}$ are coefficients.

Unfortunately, we observe that \textit{such representer theorem cannot be directly adapt to our detection problem}, as the dimensionality of $\boldsymbol{\alpha} \in \mathbb{R}^N$ scales with sample size $N$. Consequently, with significantly enlarging $N$~(very often in realistic analysis), the computational tractability of the designed detection method~(i.e., the complexity) by checking two representers of the regression coefficients $\hat{g}(\cdot) \in \mathcal{R}^{N}$ becomes prohibitively infeasible. 

To fill this gap, we establish a novel variant of the Representer Theorem operating in a subspace with fixed and finite dimensional $P < N$~($P$ is the feature size):

\begin{theorem}[Variant of Representer Theorem]\label{lem:representer_variant}
Let $X \in \mathbb{R}^{N \times d}$ be the design matrix with rank $r < N$, and let $\{\phi_i\}_{i=1}^P$ ($r \leq P < N$) form a basis for a $P$-dimensional subspace $\mathcal{S}_P \subseteq \mathcal{H}_k$ (RKHS). Consider the regularized empirical risk minimization problem:
\begin{align*}
\hat{g}(\cdot) = \underset{{g \in \mathcal{S}_P}}{\operatorname{argmin}} \left( \frac{1}{N} \sum_{i=1}^N \ell(y_i, g(\boldsymbol{x}_i)) + \xi \Omega(\|g\|_{\mathcal{H}_k}) \right),
\end{align*}
with convex loss $\ell$, convex non-decreasing regularizer $\Omega$, and $\xi > 0$. Then the minimizer $\hat{g}$ admits the representation: 
\begin{align}
\hat{g}(\cdot) = \sum_{i=1}^P \alpha_i \phi_i(\cdot), 
\end{align}
and the coefficient vector $\boldsymbol{\alpha} \in \mathbb{R}^P$ fully encodes the solution based on the sample.
\end{theorem}

\begin{remark}
    This theorem~(see detailed proof in Appendix A.1) allows us to recast the infinite-dimensional functional estimation problems for $g^{OLS}$ and $g^{HOLS}$ into fixed, finite-dimensional optimization over $\boldsymbol{\alpha} \in \mathbb{R}^P$.
\end{remark}

\subsection{Theoretical Foundation}
Leveraging Theorem \ref{lem:representer_variant}, we formulate the Kernelized Least Squares (KLS) and Higher-order Kernelized Least Squares (HKLS) estimators within a $P$-dimensional subspace $\mathcal{S}_P$ spanned by basis $\{\phi_i\}_{i=1}^P$ selected from the eigenfunctions of a symmetric positive definite kernel. Consequently, the functional regression results in~\eqref{equ:main:g1} and~\eqref{equ:main:g2} can be reduced to the $P$-dimensional $\boldsymbol{\alpha}_Z^{KLS}$ and $\boldsymbol{\alpha}_Z^{HKLS}$ solving the following minimization problems: 
\begin{enumerate}
    \item \textbf{Kernelized Least Squares (KLS)} minimizes the expected squared error loss:
    \begin{equation}
        \boldsymbol{\alpha}_Z^{KLS} = \underset{\boldsymbol{\alpha} \in \mathbb{R}^P}{\operatorname{argmin}} \, \mathbb{E}\left[ \left\| Y - \sum_{i=1}^{P} \alpha_i \phi_i(Z) \right\|^2  \right]. 
        \label{equ:alpha1}
    \end{equation}
    \item \textbf{Higher-order Kernelized Least Squares (HKLS)} minimizes the expected $\|Z\|^2$-weighted squared error loss:
    \begin{equation}
        \boldsymbol{\alpha}_Z^{HKLS} = \underset{\boldsymbol{\alpha} \in \mathbb{R}^P}{\operatorname{argmin}} \, \mathbb{E}\left[ \left\| Y - \sum_{i=1}^{P} \alpha_i \phi_i(Z) \right\|^2 \|Z\| ^2 \right].
        \label{equ:alpha2}
    \end{equation}
\end{enumerate}

\noindent {\bf Core Principle of our KRCD Method.} Based on such re-formulations, we are now ready to detect the non-linear, unobserved confounding behind the observations, by checking and comparing $\boldsymbol{\alpha}_Z^{KLS}$ and $\boldsymbol{\alpha}_Z^{HKLS}$ in each coordinate. To be specific, denoting the $i$-th coordinate of $\boldsymbol{\alpha}_Z^{KLS}$ as $\boldsymbol{\alpha}_Z^{KLS}[i]$, we prove that the coordinate-wise equality between $\boldsymbol{\alpha}_Z^{KLS}[i]$ and $\boldsymbol{\alpha}_Z^{HKLS}[i]$ is sufficient to inform the existence of unobserved confounding:

\begin{theorem}\label{thm:equivalence}
Under the causal model defined by Equations \eqref{eq1} and \eqref{eq2}, the following holds regarding the equivalence of the population estimators:
\begin{align*}
H_0:& \forall i \in [P], \quad \boldsymbol{\alpha}_Z^{KLS}[i] = \boldsymbol{\alpha}_Z^{HKLS}[i] \\
&\text{(No unobserved confounder $U$)} \\
H_1:& \exists i \in [P], \quad \boldsymbol{\alpha}_Z^{KLS}[i] \neq \boldsymbol{\alpha}_Z^{HKLS}[i] \\
&\text{(Unobserved confounder $U$ present)}
\end{align*}
\end{theorem}

Proofs of establishing equality under $H_0$ and inequality under $H_1$ are provided in Appendix A.2. In summary, this core theorem establishes the rigorous population-level theoretical foundation for KRCD: the systematic difference between $\boldsymbol{\alpha}_Z^{KLS}$ and $\boldsymbol{\alpha}_Z^{HKLS}$ serves as a valid statistical indicator of unobserved confounding presence.

\subsection{Hypothesis Testing with Finite Samples}
Although Theorem \ref{thm:equivalence} provides theoretical guidance for unobserved confounding detection by comparing regression representers $\boldsymbol{\alpha}_Z^{HKLS}$ and $\boldsymbol{\alpha}_Z^{KLS}$, a significant obstacle remains for finite-sample algorithmic implementation. 

Specifically, the minimization problems defined in~\eqref{equ:alpha1} and~\eqref{equ:alpha2} assume infinite samples (i.e., population expectations $\mathbb{E}$), while finite-sample estimates of $\boldsymbol{\alpha}_Z^{HKLS}$ and $\boldsymbol{\alpha}_Z^{KLS}$ may deviate from their population counterparts. Consequently, direct comparison of these coefficients with finite samples risks incorrect detection.

To overcome this issue, we first derive the closed-form solutions of empirical estimators $\widehat{\boldsymbol{\alpha}_Z^{KLS}}$ and $\widehat{\boldsymbol{\alpha}_Z^{HKLS}}$ based on empirical samples $\{(Z_j, Y_j)\}_{j=1}^N$. With the kernel basis $k(Z_i, \cdot)$ (setting $\phi_i(\cdot) = k(Z_i, \cdot)$, $\Omega(\|g\|_{\mathcal{H}_k}) = \sum_{i=1}^{P} \alpha_i^2$, and $P < N$), then we have:

\begin{theorem}\label{thm:finite_estimators}
The empirical KLS and HKLS estimators are given by:
\begin{equation}
\begin{aligned}
\widehat{\boldsymbol{\alpha}_Z^{KLS}} &= (K_{ZZ} K_{ZZ}^\top + N \xi I_P)^{-1} K_{ZZ} \boldsymbol{Y}, \\
\widehat{\boldsymbol{\alpha}_Z^{HKLS}} &= (K_{ZZ} \Psi K_{ZZ}^\top  + N \xi I_P)^{-1} K_{ZZ} \Psi \boldsymbol{Y},
\end{aligned}
\end{equation}
where $K_{ZZ} \in \mathbb{R}^{P \times N}$ with $[K_{ZZ}]_{ij} = k(Z_i, Z_j)$, $\boldsymbol{Y} = [Y_1,\dots,Y_N]^\top$, and $\Psi = \operatorname{diag}(\|Z_1\|^2,\dots,\|Z_N\|^2)$.
\end{theorem}

As mentioned before, diverse risk factors, e.g., sampling variability and outcome noise, will lead to perturbation of the difference vector $\hat{\boldsymbol{\delta}} = \widehat{\boldsymbol{\alpha}_Z^{HKLS}} - \widehat{\boldsymbol{\alpha}_Z^{KLS}}$, served as the diagnostic marker for unobserved confounding. We therefore present the last theorem in below to indicate the asymptotic normality of $\hat{\boldsymbol{\delta}}$, which further guides the construction of the hypothesis testing with finite samples:

\begin{theorem}[Asymptotic Distribution of $\hat{\boldsymbol{\delta}}$]\label{thm:asym_dist}
Assume standard regularity conditions (specified in Appendix A.3) hold, and the causal model \eqref{eq1}, \eqref{eq2} is correct. Under $H_0$ ($\boldsymbol{\alpha}^{KLS} = \boldsymbol{\alpha}^{HKLS}$), as $N \to \infty$:
\begin{equation}
\sqrt{N} \hat{\boldsymbol{\delta}} = \sqrt{N} (\widehat{\boldsymbol{\alpha}_Z^{HKLS}} - \widehat{\boldsymbol{\alpha}_Z^{KLS}}) \xrightarrow{D} \mathcal{N} (\boldsymbol{0}, \sigma^2 \Gamma), \label{eq:asym_dist}
\end{equation}
where the asymptotic covariance matrix $\sigma^2 \Gamma$ is consistently estimable. 
\end{theorem}

This asymptotic characterization enables practical hypothesis testing for $H_0$. Specifically, the established asymptotic normality allows constructing component-wise standardized tests whose null distributions are approximately standard normal. Accounting for multiple testing via Bonferroni correction, the null hypothesis of no unobserved confounding is rejected if any standardized component difference shows statistically significant deviation at the adjusted significance level.

\subsection{Complexity Analysis}
We derive the overall complexity of our proposed KRCD method as $O\left(N^{2} d + P^{3} + P^{2} N\right)$, in terms of the sample size $N$, parameter dimension $P$, and feature dimension $d$ (detailed derivation provided in Appendix A.4).

\begin{algorithm}[H]
\caption{Kernel Regression Confounder Detection}
\label{alg:simplified_KRCD}
\textbf{Input}: Observed covariates $X \in \mathbb{R}^{N \times d}$, treatment $T \in \mathbb{R}^N$, outcome $Y \in \mathbb{R}^N$, parameter dimension $P$, regularization parameter $\lambda = N \xi$. \\
\textbf{Output}: Detect conclusion
\begin{algorithmic}[1]
    \STATE $Z \gets [T, X]$
    \STATE $K \gets \text{rows}_{1:P}(\text{Kernel Function}(Z))$ 
    \STATE $K_\psi \gets K \odot (\text{diag}(Z Z^\top))^\top$
    \STATE $\alpha_{KLS} \gets \left(K K^\top + \lambda I\right)^{-1} K Y$
    \STATE $\alpha_{HKLS} \gets \left(K_\psi K^\top + \lambda I\right)^{-1} (K_\psi Y)$
    \STATE $V_0 \gets (K_\psi K^\top + \lambda I)^{-1} K_\psi - (K K^\top + \lambda I)^{-1} K$
    \STATE $V \gets N \cdot (V_0 V_0^\top)$
    \STATE $\sigma^2 \gets \dfrac{1}{N} (Y - K^\top \alpha_{KLS})^\top (Y - K^\top \alpha_{KLS})$
    \FOR{$j = 1$ \TO $P$}
        \STATE $p_j \gets 2(1 - \Phi(|\dfrac{\sqrt{N}(\alpha_{HKLS,j} - \alpha_{KLS,j})}{\sqrt{\sigma^2 V_{jj}}}|))$
    \ENDFOR
    \STATE Apply Bonferroni correction: $\delta_j \gets \mathds{1}[p_j < 0.05/P]$  
    \IF{$\sum \delta_j = 0$}
        \STATE \textbf{return} "Support null hypothesis"
    \ELSE
        \STATE \textbf{return} "Reject null hypothesis"
    \ENDIF
\end{algorithmic}
\end{algorithm}

\section{Experiments}
\subsubsection{Benchmarks.}
To comprehensively evaluate KRCD's capability for unobserved confounding detection, we conducted simulation studies using both synthetic datasets and the Twins dataset \citep{almond2005costs}. Unless explicitly stated otherwise, all experiments employed non-Gaussian independent noise components and a default regularization parameter of $\lambda = 10^{-8}$. Each parameter combination was repeated 30 times with sample sizes of 1000 observations, and statistical significance was assessed at the conventional $\alpha = 0.05$ threshold.
\subsubsection{Baselines.}
This paper focuses on state-of-the-art (SOTA) confounding detection methods for comparative examination, including: ROCD~\citep{schultheiss2024higher}, ME-ICM~\citep{karlsson2023detecting}, and CNF~\citep{reddy2024detecting}. (See details of baselines in Appendix B)
\subsubsection{Evaluation Protocols.}
We evaluate the proposed KRCD in a variety of scenarios, including~(see details of scenarios in Appendix C):
\begin{itemize}
\item \textit{Nonlinear Single-Environment.} All data is generated by identical nonlinear structural causal mechanisms under a fixed causal model.
\item \textit{Nonlinear Multi-Environment.} Distinct environments exhibit different nonlinear causal mechanisms within a shared causal framework.
\end{itemize}
The evaluation metrics including:
\begin{itemize}
\item \textit{ROC Curve \& AUC.} Sensitivity and False Positive Rate over thresholds; AUC quantifies overall performance of this trade-off.
\item \textit{Detection Rate.} The method's detection rate for unobserved confounders, quantifying sensitivity (with confounders) and false positive rate (without confounders).
\item \textit{Runtime.} Total time required to complete all detection runs at a fixed confounding strength, benchmarking computational efficiency.
\end{itemize}
\subsubsection{Questions.}
The empirical evaluation addresses the following three research questions:
\begin{itemize}
\item Whether the proposed KRCD method outperforms detection baselines in our NSE problem?
\item Whether the proposed KRCD method outperforms detection baselines in multi-environment settings?
\item Can KRCD achieve superior computational efficiency in detecting hidden confounding, outperforming baselines?
\item Can KRCD exhibit robustness when the grounding structural function, fail to fall inside the kernel space?
\item How sensitive is KRCD's performance to changes in the regularization parameter?
\end{itemize}

\subsection{Performance in Our NSE Setup}
\begin{figure*}[t]
  \centering
  \begin{subfigure}{0.32\textwidth}
    \centering
    \includegraphics[width=\linewidth]{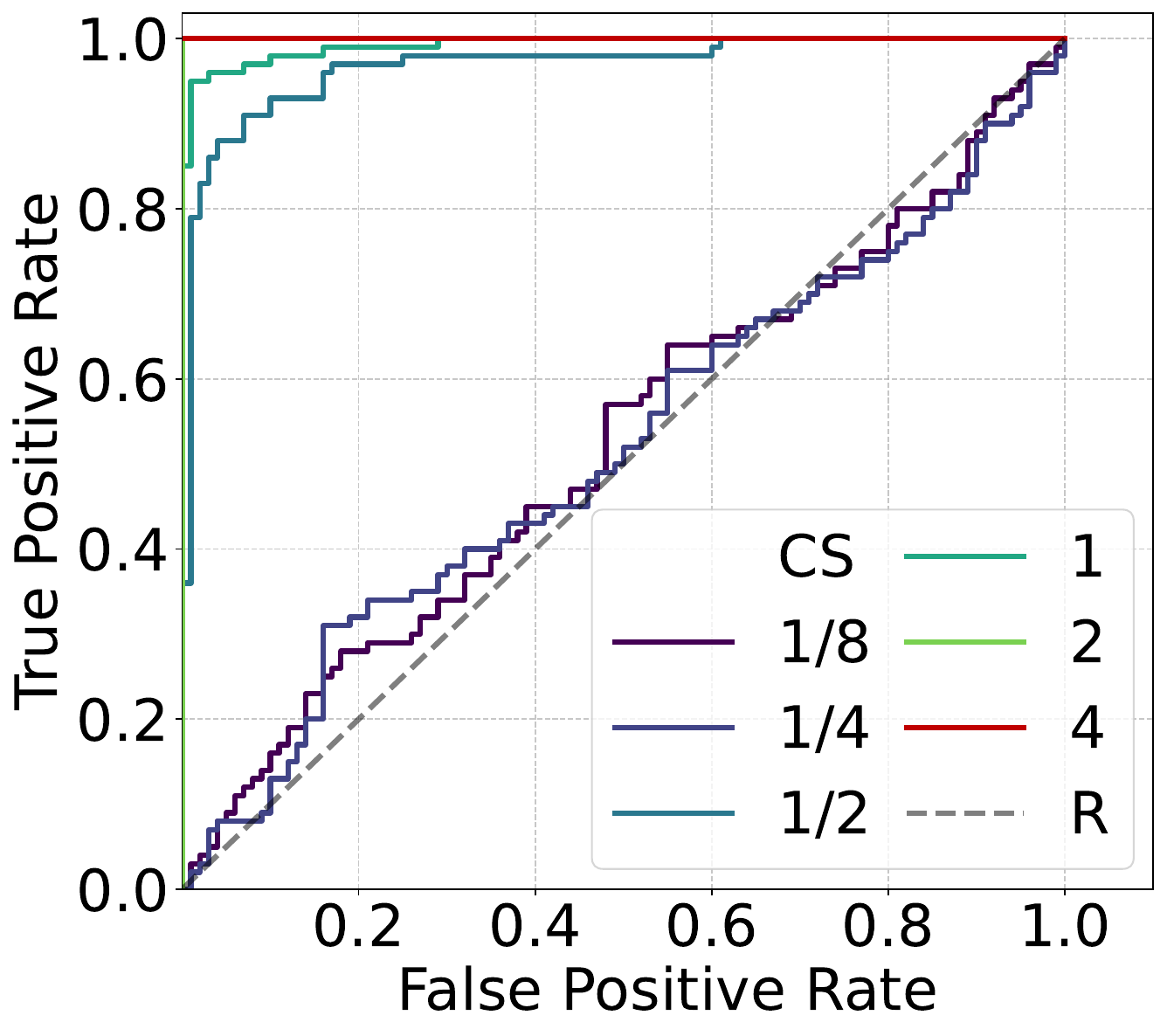}
    \caption{}
    \label{fig2-1}
  \end{subfigure}
  \hfill
  \begin{subfigure}{0.32\textwidth}
    \centering
    \includegraphics[width=\linewidth]{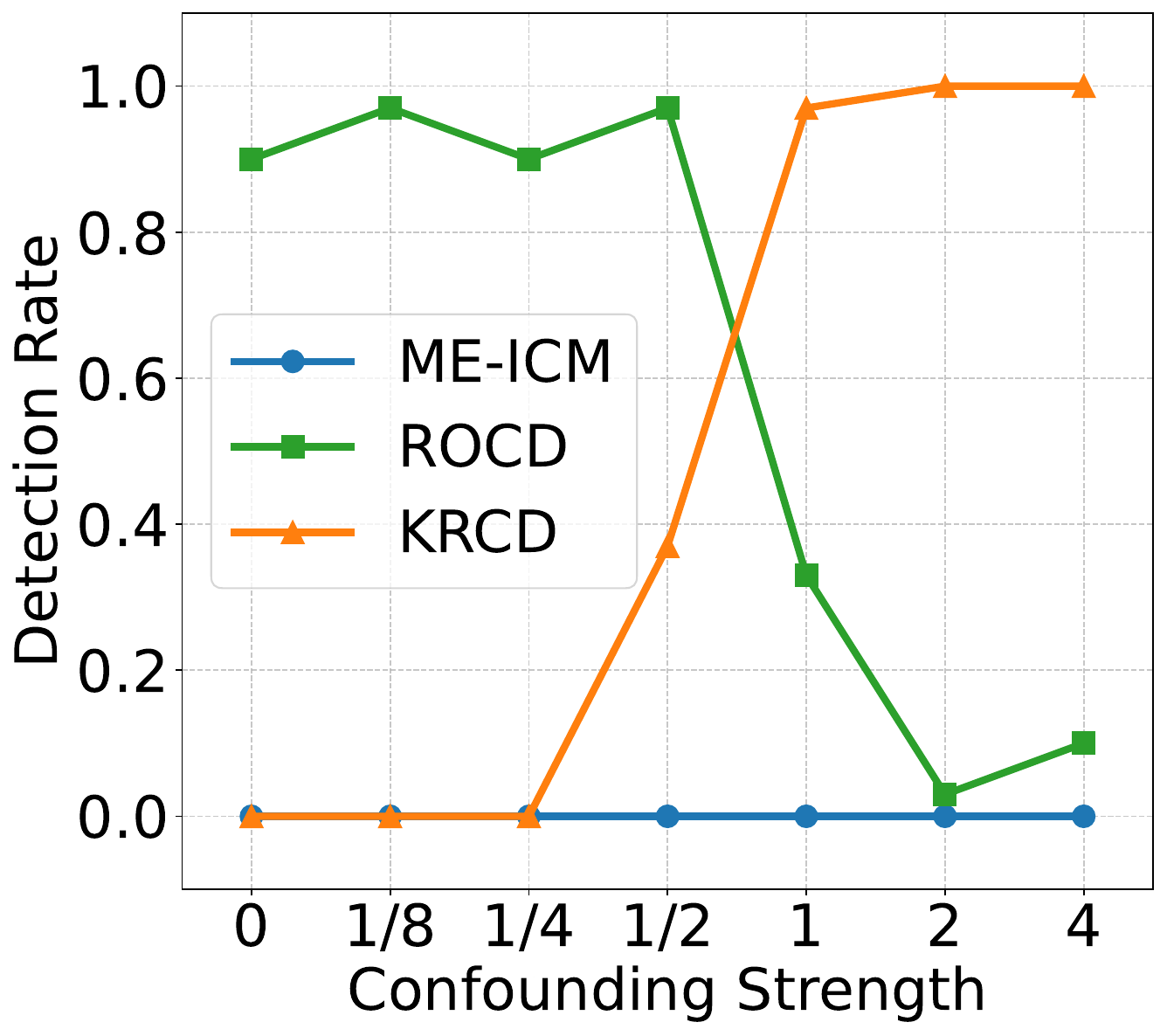}
    \caption{}
    \label{fig2-2}
  \end{subfigure}
  \hfill
  \begin{subfigure}{0.32\textwidth}
    \centering
    \includegraphics[width=\linewidth]{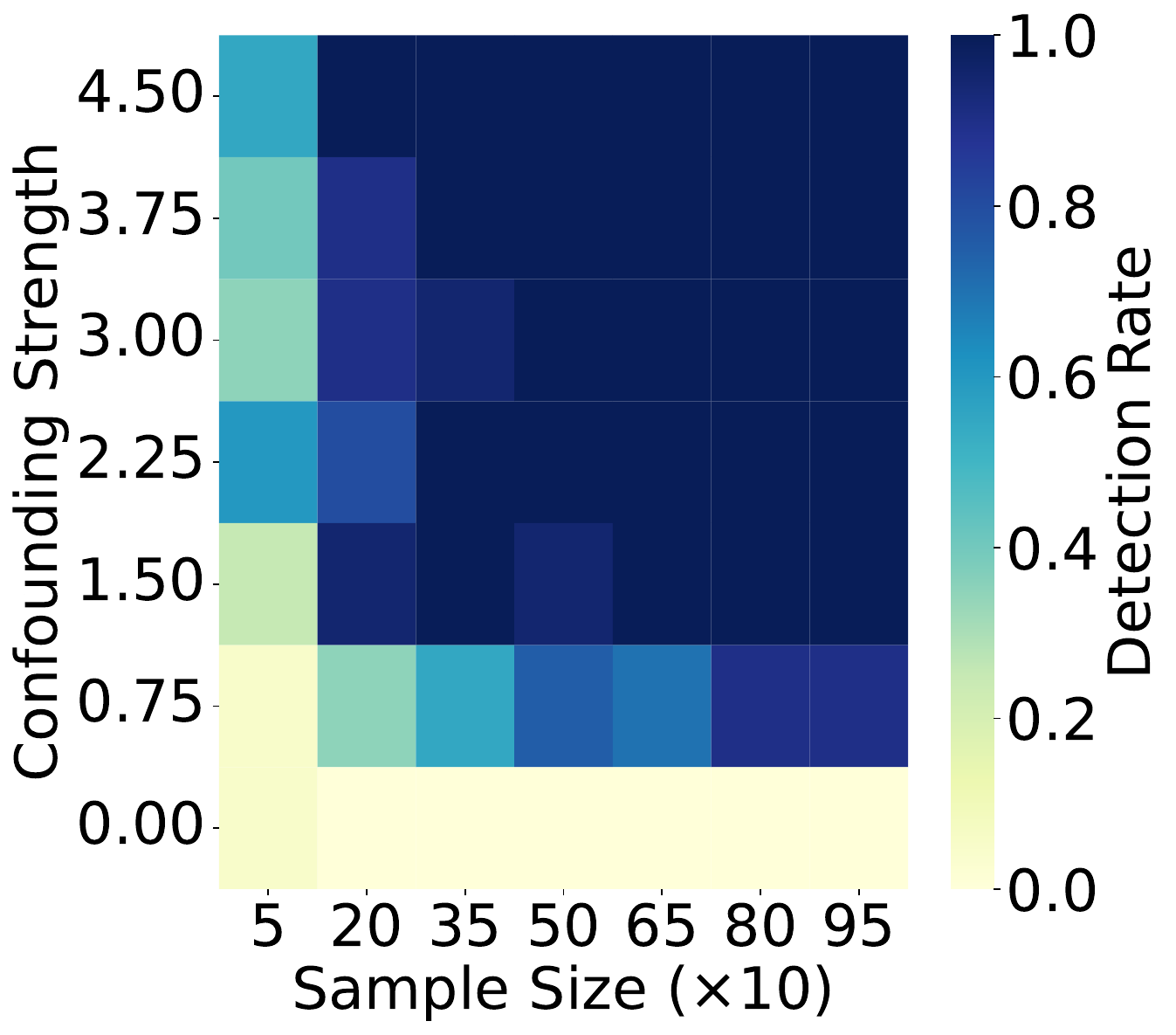}
    \caption{}
    \label{fig2-3}
  \end{subfigure}
  \caption{KRCD Performance in Nonlinear Single-Environment---(a) ROC Curves across Confounding Strengths. (b) Detection Rate of KRCD, ROCD, and ME-ICM. (c) Detection Rate across Confounding Strengths and Sample Sizes.}
  \label{fig2}
\end{figure*}
To validate KRCD's performance in our NSE setup, we performed systematic experiments on the Twins dataset where all environments share identical causal mechanisms.

As shown in Figure~\ref{fig2-1}, KRCD remains robust under null confounding and shows high sensitivity once confounding strength reaches a threshold ($\geq 0.5$). Figure~\ref{fig2-3} indicates that KRCD detects unobserved confounding with only 200 samples, highlighting its sample efficiency.

Figure~\ref{fig2-2} compares single-environment nonlinear confounding detection. ME-ICM, requiring multiple environments, fails completely here. ROCD, based on linearity assumptions, misinterprets nonlinearity as confounding, yielding high false positives under null confounding and declining detection under strong confounding as true confounders mask nonlinear features. In contrast, KRCD maintains $0\%$ detection under null confounding (controlling Type~I error) and exceeds $95\%$ detection at confounding strengths $\geq 1$.

These results confirm KRCD's strong single-environment performance: robustness to null confounding, high sensitivity beyond a strength threshold, low sample requirements, and clear advantages over methods needing linear assumptions or multi-environment data.

\subsection{Performance in Multi-Environment}

\begin{figure*}[t]
  \centering
  \begin{subfigure}{0.32\textwidth}
    \centering
    \includegraphics[width=\linewidth]{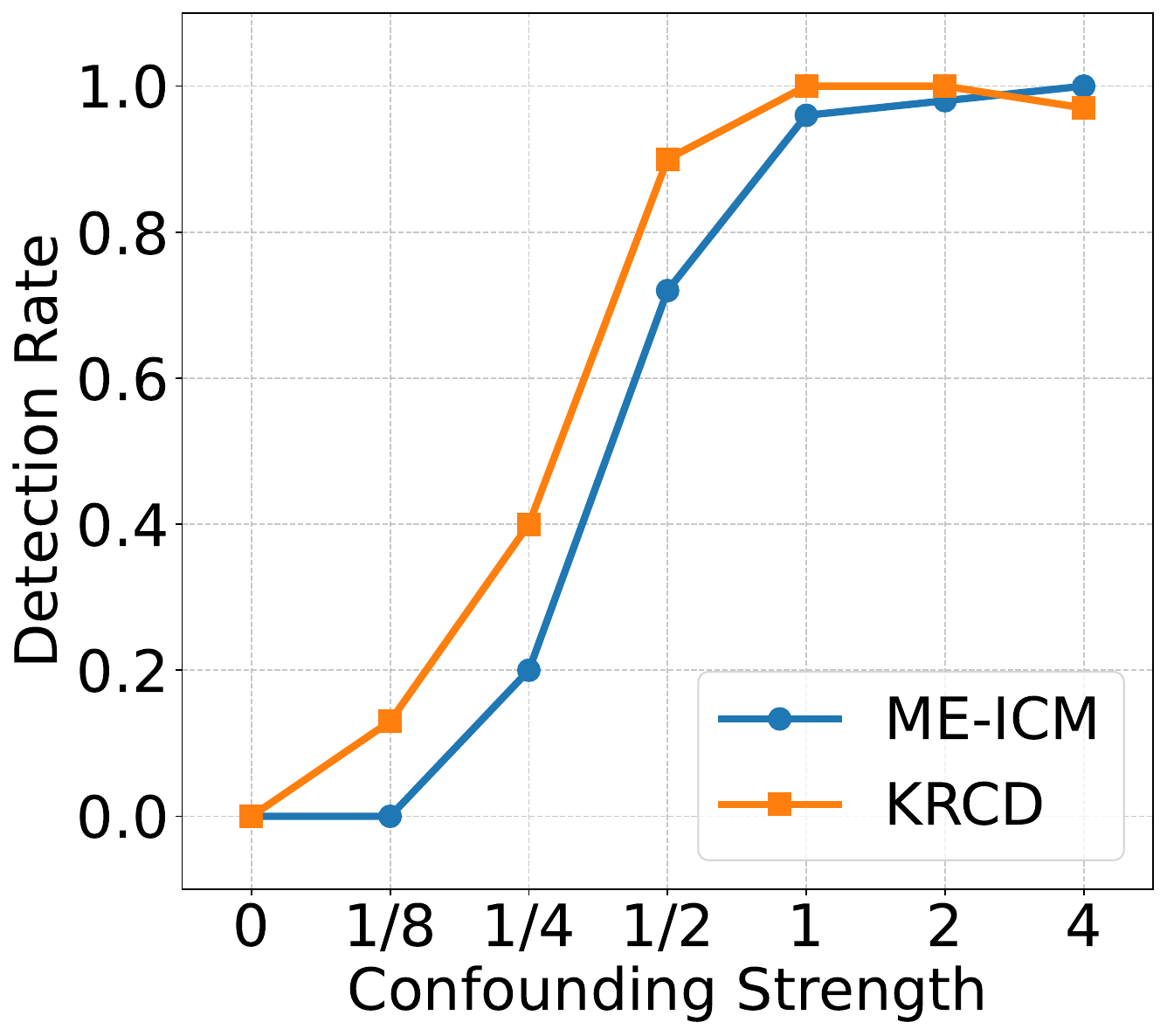}
    \caption{}
    \label{fig3-1}
  \end{subfigure}
  \hfill
  \begin{subfigure}{0.32\textwidth}
    \centering
    \includegraphics[width=\linewidth]{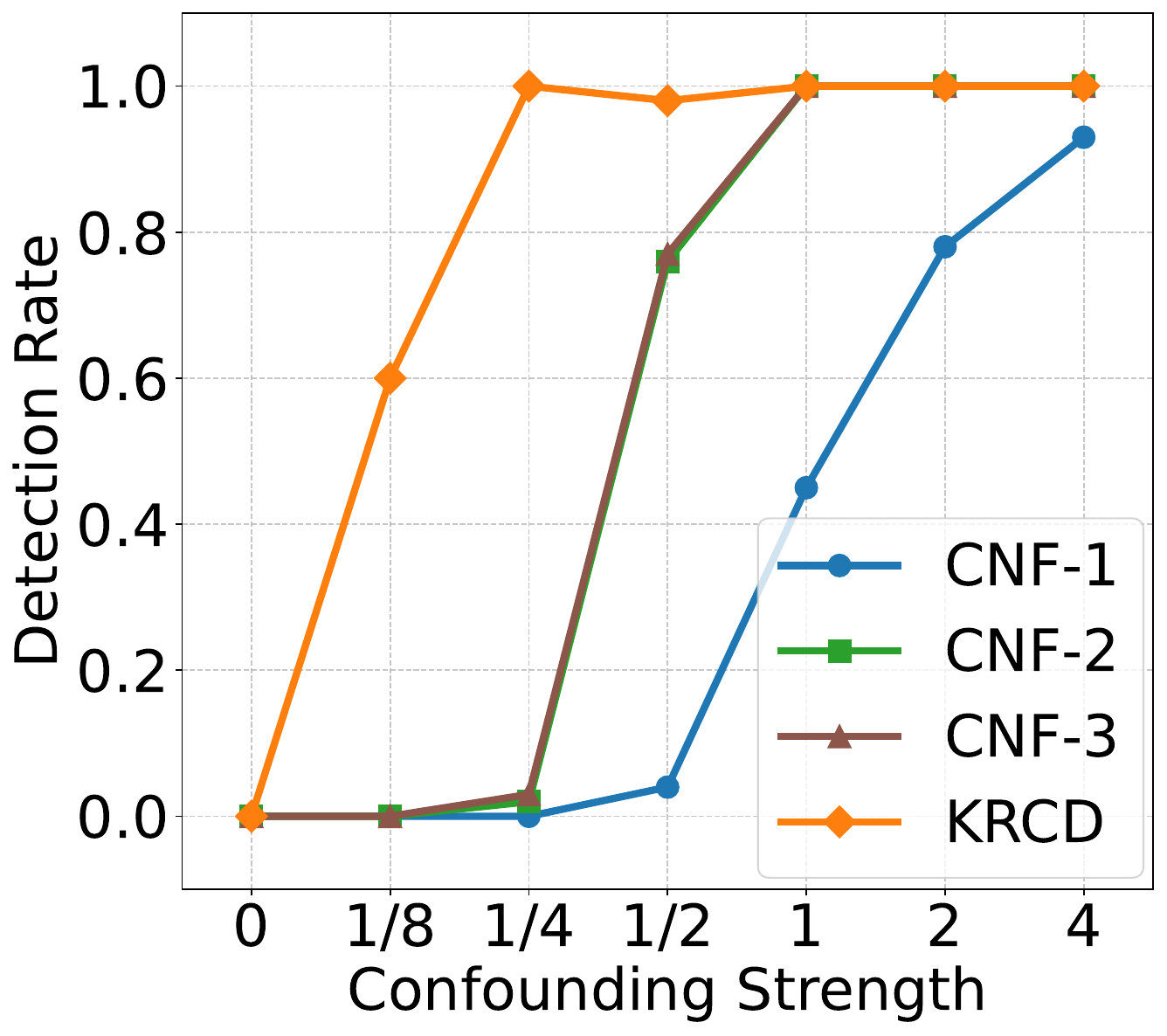}
    \caption{}
    \label{fig3-2}
  \end{subfigure}
  \hfill
  \begin{subfigure}{0.32\textwidth}
    \centering
    \includegraphics[width=\linewidth]{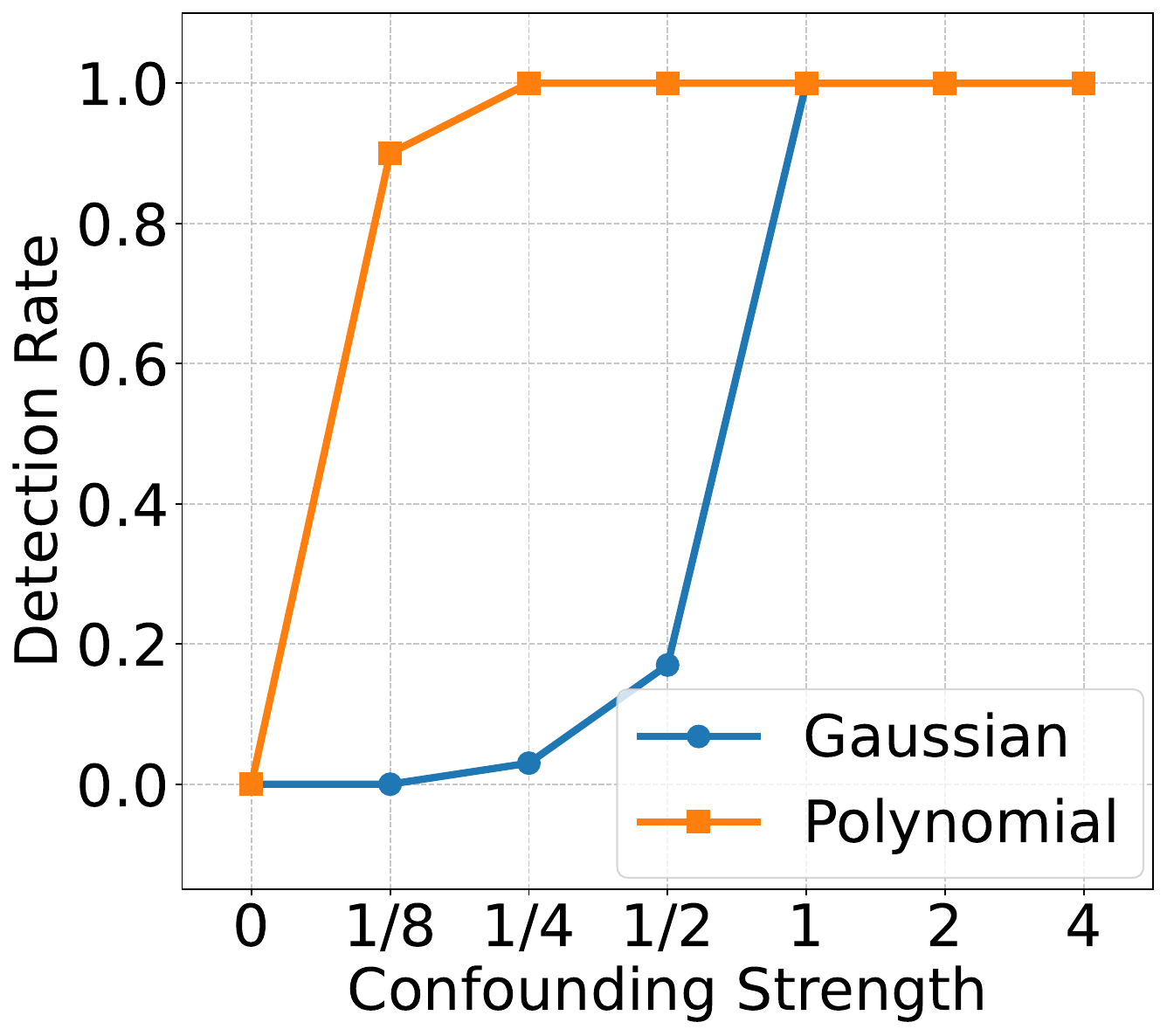}
    \caption{}
    \label{fig3-3}
  \end{subfigure}
  \caption{(a) Detection Rate of KRCD and ME-ICM in Nonlinear Multi-Environment. (b)  Detection Rate of KRCD and CNF in Nonlinear Multi-Environment. (c) Performance of Polynomial-KRCD and Gaussian-KRCD.}
  \label{fig3}
\end{figure*}

To validate KRCD's performance in multi-environment settings, we conducted systematic experiments where environments exhibit distribution shifts in causal mechanisms while maintaining consistent causal structures. Due to divergent environmental definitions across baselines, evaluations were performed separately on Twins and synthetic datasets.

\subsubsection{Comparison with ME-ICM.}

On the Twins dataset, KRCD fundamentally differs by integrating environmental variables as covariates rather than using ME-ICM's explicit partitioning. Figure~\ref{fig3-1} shows both methods achieve comparable performance, demonstrating KRCD's competitive multi-environment adaptability.

\subsubsection{Comparison with CNF.}

On the synthetic dataset, KRCD outperforms CNF in detecting unobserved confounders, particularly under weak confounding, as evidenced by Figure~\ref{fig3-2}. This performance gap occurs because CNF's blocking approach, while computationally efficient, loses information and reduces sensitivity to weaker confounders.

\subsection{Computational Efficiency Analysis}
To evaluate KRCD's computational efficiency, we compared runtime of all baselines in experiments for Figure \ref{fig3-1} and \ref{fig3-2}, defined as the time for 100 trials. Table \ref{tab1} shows that on the Twins dataset, KRCD is substantially faster than ME-ICM, markedly reducing computational time. On the Synthetic dataset, KRCD has efficiency similar to CNF.

Notably, KRCD's significant efficiency variation across datasets stems directly from its parameter dimension $P$: Setting $P=999$ on the Twins dataset prioritizes modeling precision, while $P=40$ on the Synthetic dataset maximizes computational efficiency. This tunability enables KRCD to flexibly balance predictive performance against computational demands through $P$ adjustment, ensuring adaptability to diverse resource constraints.

\begin{table}
\centering
\setlength{\tabcolsep}{8pt}
\begin{tabular}{lcc}
\toprule
\textbf{Dataset} & \textbf{Method} & \textbf{Runtime(s/it)} \\
\midrule
\multirow{2}{*}{Twins} 
    & KRCD & 68.748 \\
    & ME-ICM & 2657.838 \\
\midrule
\multirow{4}{*}{Synthetic} 
    & KRCD & 0.184 \\
    & CNF-1 & 0.072 \\
    & CNF-2 & 0.048 \\
    & CNF-3 & 0.046 \\
\bottomrule
\end{tabular}
\caption{Comparison of Runtime for KRCD and Baselines}
\label{tab1}
\end{table}

\subsection{Robustness under Kernel Misspecification}
To evaluate the computational efficiency of our KRCD method, we compared runtime across all baselines in the experiments for Figure \ref{fig3-1} and \ref{fig3-2}, defined as the time for 100 trials. Table \ref{tab1} shows that on the Twins dataset, KRCD is substantially faster than ME-ICM, markedly reducing computational time. On the Synthetic dataset, KRCD has efficiency similar to CNF baselines.

As demonstrated in Figure~\ref{fig3-3}, both kernel configurations of KRCD maintain strict Type I error control (no false positives). Polynomial-KRCD achieves high sensitivity even with minimal confounding conditions, maintaining full detection as confounding strength increases. Gaussian-KRCD needs stronger signals but achieves full detection at high confounding strength. KRCD thus provides robust, reliably detects unobserved confounding without false alarms and excels in high-confounding scenarios.

\iffalse
\begin{figure}[t]
  \centering
  \includegraphics[width=0.8\columnwidth]{KRCD_05.pdf}
  \caption{Performance of Polynomial-KRCD and Gaussian-KRCD.}
  \label{fig5}
\end{figure}
\fi

\subsection{Parameter Sensitivity Analysis}
The regularization parameter $\lambda$ in Algorithm \ref{alg:simplified_KRCD} prevents overfitting and ensures numerical stability during kernel matrix inversion. As Table \ref{tab2} shows, KRCD exhibits minimal sensitivity to $\lambda$ across wide parameter ranges ($10^{-12}$ to $10^{-4}$), maintaining stable AUC performance. Within this broad operational window, the method consistently achieves high accuracy ($\geq 0.95$ AUC for confounding strengths $\geq 0.5$), demonstrating significant robustness to parameter selection. Performance degradation occurs only under extreme regularization ($\lambda=1$), where over-smoothing obscures essential confounding signals---particularly detrimental at medium confounding strength with clinically significant effects. This resilience to $\lambda$ variation confirms KRCD's practical usability in diverse settings.

\begin{table}
\centering
\setlength{\tabcolsep}{8pt}
\begin{tabular}{c|cccc}
\toprule
\multirow{2}{*}{\textbf{Conf Strength}} & \multicolumn{4}{c}{\textbf{Parameter $\lambda$}} \\
& $10^{-12}$ & $10^{-8}$ & $10^{-4}$ & $1$ \\
\midrule
0.25 & 0.468 & 0.507 & 0.473 & 0.404 \\
0.5  & 0.959 & 0.980 & 0.981 & 0.257 \\
1  & 0.998 & 1.000 & 1.000 & 0.720 \\
2   & 1.000 & 1.000 & 1.000 & 1.000 \\
\bottomrule
\end{tabular}
\caption{KRCD AUC Performance Sensitivity to $\lambda$}
\label{tab2}
\end{table}

\section{Conclusion}
This paper proposes the Kernel Regression Confounder Detection (KRCD) method, which can detect unobserved confounders in nonlinear single-environment observational data, addressing the limitations of existing methods. The core principle of KRCD involves using kernel regression within reproducing kernel Hilbert spaces to model complex relationships; it compares the results of standard kernel regression and higher-order kernel regression to construct a test statistic--a significant deviation of this statistic from zero indicates the presence of unobserved confounding. Experiments on synthetic benchmark datasets and the Twins dataset demonstrate that KRCD not only significantly outperforms baseline methods in detection performance but also achieves superior computational efficiency.

While KRCD effectively detects unobserved confounding in nonlinear single-environment settings, it cannot quantify confounding strength. A promising direction is developing a novel framework for quantifying confounding strength based on KRCD methodology.

\section*{Ethics Statement}
This work does not raise any ethical concerns. 
All experiments are conducted on publicly available datasets, and no human subjects or sensitive attributes are involved.

\section*{Acknowledgments}
This work is supported in part by the National Natural Science Foundation of China under Grants No. 62372459, 62525213, and 62376243, in part by the NUDT Youth Independent Innovation Science Fund under Grant No. ZK25-20, in part by the "Pioneer" and "Leading Goose" R\&D Program of Zhejiang (2025C02037).

\bibliography{KRCD}

\appendix
\section{Theoretical Proofs}
\subsection{Variant of Representer Theorem}
\begin{theorem}[Variant of Representer Theorem]\label{lem:representer_variant}
Let $X \in \mathbb{R}^{N \times d}$ be the design matrix with rank $r < N$, and let $\{\phi_i\}_{i=1}^P$ ($r \leq P < N$) form a basis for a $P$-dimensional subspace $\mathcal{S}_P \subseteq \mathcal{H}_k$ (RKHS). Consider the regularized empirical risk minimization problem:
\begin{align*}
\hat{g}(\cdot) = \underset{{g \in \mathcal{S}_P}}{\operatorname{argmin}} \left( \frac{1}{N} \sum_{i=1}^N \ell(y_i, g(\boldsymbol{x}_i)) + \xi \Omega(\|g\|_{\mathcal{H}_k}) \right),
\end{align*}
with convex loss $\ell$, convex non-decreasing regularizer $\Omega$, and $\xi > 0$. Then the minimizer $\hat{g}$ admits the representation: 
\begin{align*}
\hat{g}(\cdot) = \sum_{i=1}^P \alpha_i \phi_i(\cdot), 
\end{align*}
and the coefficient vector $\boldsymbol{\alpha} \in \mathbb{R}^P$ fully encodes the solution based on the sample.
\end{theorem}
\begin{Proof}
The proof consists of two main parts:

\noindent \textbf{Part 1: Existence and Uniqueness}
\begin{enumerate}
    \item \textit{Finite-dimensional parameterization}: 
    Since $\mathcal{S}_P$ is a $P$-dimensional subspace with basis $\{\phi_i\}_{i=1}^P$, every $g \in \mathcal{S}_P$ admits representation $g = \sum_{i=1}^P \alpha_i \phi_i$. Thus:
    \begin{align*}
    \min_{g \in \mathcal{S}_P} J(g) \equiv \min_{\boldsymbol{\alpha} \in \mathbb{R}^P} \bigg\{ &\frac{1}{N} \sum_{i=1}^N \ell\left(y_i, \left(\textstyle{\sum_{j=1}^P \alpha_j \phi_j}\right)(\boldsymbol{x}_i)\right) \\
    &+ \xi \Omega\left( \left\| \textstyle{\sum_{j=1}^P \alpha_j \phi_j} \right\|_{\mathcal{H}_k} \right) \bigg\}
    \end{align*}
    
    \item \textit{Convexity and existence}:
    \begin{itemize}
        \item $\ell(y_i, \cdot)$ convex and $f_i(\boldsymbol{\alpha}) = \sum_{j=1}^P \alpha_j \phi_j(\boldsymbol{x}_i)$ linear $\implies$ convex in $\boldsymbol{\alpha}$
        \item Regularizer: $\boldsymbol{\alpha} \mapsto \| \sum_{j=1}^P \alpha_j \phi_j \|_{\mathcal{H}_k}$ is a norm (by basis linear independence), hence convex. Composition with convex non-decreasing $\Omega$ preserves convexity
        \item $J(\boldsymbol{\alpha})$ convex as sum of convex functions
        \item Coercivity: $\|\boldsymbol{\alpha}\| \to \infty \implies \Omega(\|g\|_{\mathcal{H}_k}) \to \infty$
    \end{itemize}
    By convex analysis, a minimizer $\boldsymbol{\alpha}^*$ exists.
    
    \item \textit{Uniqueness}:
    \begin{itemize}
        \item If $\ell(y,\cdot)$ strictly convex: $\ell(y_i, f_i(\boldsymbol{\alpha}))$ strictly convex (linear injective $f_i$)
        \item If $\Omega$ strictly convex: $\Omega(\|g\|_{\mathcal{H}_k})$ strictly convex in $\boldsymbol{\alpha}$ (linear bijective map)
    \end{itemize}
    In either case, $J$ strictly convex $\implies$ unique $\boldsymbol{\alpha}^*$.
\end{enumerate}

\noindent \textbf{Part 2: Fitting Capability}

Let $\hat{g}_{\text{class}} = \sum_{k=1}^N \beta_k k(\boldsymbol{x}_k, \cdot)$ be the classical Representer Theorem solution. Since $\mathrm{span}\{\phi_i\}_{i=1}^P \supseteq \mathrm{Col}(X)$ (as $\mathrm{rank}(X) = r \leq P$), for each $\boldsymbol{x}_i$:
\begin{align*}
k(\boldsymbol{x}_i, \cdot) = \sum_{j=1}^P \gamma_{ij} \phi_j \quad \text{(data embedding in $\mathcal{S}_P$)}
\end{align*}
Thus $\hat{g}_{\text{class}}$ has equivalent representation in $\mathcal{S}_P$:
\begin{align*}
\hat{g}_{\text{class}} = \sum_{k=1}^N \beta_k k(\boldsymbol{x}_k, \cdot) = \sum_{j=1}^P \left( \sum_{k=1}^N \beta_k \gamma_{kj} \right) \phi_j = \sum_{j=1}^P \tilde{\alpha}_j \phi_j
\end{align*}
Since $\hat{g}_{\text{class}}$ is optimal in $\mathcal{H}_k$ and $\hat{g}_{\text{class}} \in \mathcal{S}_P$, the constrained minimizer $\hat{g}$ must satisfy for all $i=1,\dots,N$:
\begin{align*}
\hat{g}(\boldsymbol{x}_i) = \langle \hat{g}, k(\boldsymbol{x}_i, \cdot) \rangle = \langle \hat{g}_{\text{class}}, k(\boldsymbol{x}_i, \cdot) \rangle = \hat{g}_{\text{class}}(\boldsymbol{x}_i)
\end{align*}
because any deviation would violate the optimality of $\hat{g}_{\text{class}}$ or $\hat{g}$ in their respective spaces. Thus $\boldsymbol{\alpha}^*$ fully encodes $\hat{g}$ relative to the sample.
\end{Proof}

\subsection{Unobserved Confounder Detection Equivalence}
\begin{theorem}\label{thm:equivalence}
Under the causal model, considering the null hypothesis $H_0$ and the alternative hypothesis $H_1$, the following holds regarding the equivalence of the population estimators:
\begin{align*}
H_0:& \forall i \in [P], \quad \boldsymbol{\alpha}_Z^{KLS}[i] = \boldsymbol{\alpha}_Z^{HKLS}[i] \\
&\text{(No unobserved confounder $U$)} \\
H_1:& \exists i \in [P], \quad \boldsymbol{\alpha}_Z^{KLS}[i] \neq \boldsymbol{\alpha}_Z^{HKLS}[i] \\
&\text{(Unobserved confounder $U$ present)}
\end{align*}
\end{theorem}
Building upon the variant of Representer Theorem, the structural equation under $H_0$ and $H_1$ can be restructured into the following unified mathematical formulations:
\begin{align}
H_0: Y &= \sum_{i=1}^{P} \alpha_i \phi_i(Z) + \epsilon_Y, \label{eq:A1}\\
H_1: Y &= \sum_{i=1}^{P} \alpha_i \phi_i(Z) + h_Y(U) + \epsilon_Y, \label{eq:A2}
\end{align}
with define the population regression coefficients $\boldsymbol{\alpha}_Z = [\alpha_1,\dots,\alpha_p]^\top$. 
\begin{Proof}
The proof consists of two main parts:

\noindent \textbf{Part 1: $\boldsymbol{\alpha}_Z^{KLS} = \boldsymbol{\alpha}_Z^{HKLS}$ under $H_0$}

Given structural equation \eqref{eq:A1}, the KLS and HKLS estimators take the form:
\begin{align*}
\boldsymbol{\alpha}_Z^{KLS} &= \underset{\boldsymbol{\alpha} \in \mathbb{R}^P}{\operatorname{argmin}} \, \mathbb{E}\left[ \left\| Y - \sum_{i=1}^{P} \alpha_i \phi_i(Z) \right\|^2 \right] \\
&= (A^{KLS})^{-1}\boldsymbol{b}^{KLS} \\
&\overset{H_0}{=} \boldsymbol{\alpha}_Z, \\
\boldsymbol{\alpha}_Z^{HKLS} &= \underset{\boldsymbol{\alpha} \in \mathbb{R}^P}{\operatorname{argmin}} \, \mathbb{E}\left[ \left\| Y - \sum_{i=1}^{P} \alpha_i \phi_i(Z) \right\|^2 \|Z\| ^2 \right] \\
&= (A^{HKLS})^{-1}\boldsymbol{b}^{HKLS} \\
&\overset{H_0}{=} \boldsymbol{\alpha}_Z,
\end{align*}
where matrices and vectors are defined as: 
\begin{align*}
A_{ji}^{KLS} &= \mathbb{E}\big[\phi_i(Z) \phi_j(Z)\big], \\
\boldsymbol{b}_{j}^{KLS} &= \mathbb{E}\big[Y \phi_j(Z)\big], \\
A_{ji}^{HKLS} &= \mathbb{E}\big[\phi_i(Z) \phi_j(Z)\left\| Z \right\|^2\big], \\
\boldsymbol{b}_{j}^{HKLS} &= \mathbb{E}\big[Y \phi_j(Z)\left\| Z \right\|^2\big].
\end{align*}
Therefore, under the null hypothesis $H_0$, the parameters $\boldsymbol{\alpha}_Z^{KLS}$ and $\boldsymbol{\alpha}_Z^{HKLS}$ are equal. 

\noindent \textbf{Part 2: $\boldsymbol{\alpha}_Z^{KLS} \neq \boldsymbol{\alpha}_Z^{HKLS}$ under $H_1$}

Given structural equation \eqref{eq:A2}, the KLS and HKLS estimators take the form:
\begin{align*}
\boldsymbol{\alpha}_Z^{KLS} &= \underset{\boldsymbol{\alpha} \in \mathbb{R}^P}{\operatorname{argmin}} \, \mathbb{E}\left[ \left\| Y - \sum_{i=1}^{P} \alpha_i \phi_i(Z) \right\|^2 \right] \\
&= (A^{KLS})^{-1}\boldsymbol{b}^{KLS} \\
&\overset{H_1}{=} (A^{KLS})^{-1}\boldsymbol{c}^{KLS} + \boldsymbol{\alpha}_Z, \\
\boldsymbol{\alpha}_Z^{HKLS} &= \underset{\boldsymbol{\alpha} \in \mathbb{R}^P}{\operatorname{argmin}} \, \mathbb{E}\left[ \left\| Y - \sum_{i=1}^{P} \alpha_i \phi_i(Z) \right\|^2 \|Z\| ^2 \right] \\
&= (A^{HKLS})^{-1}\boldsymbol{b}^{HKLS} \\
&\overset{H_1}{=} (A^{HKLS})^{-1}\boldsymbol{c}^{HKLS} + \boldsymbol{\alpha}_Z,
\end{align*}
where matrices and vectors are defined as: 
\begin{align*}
A_{ji}^{KLS} &= \mathbb{E}\big[\phi_i(Z) \phi_j(Z)\big], \\
\boldsymbol{b}_{j}^{KLS} &= \mathbb{E}\big[Y \phi_j(Z)\big], \\
\boldsymbol{c}_{j}^{KLS} &= \mathbb{E}\big[h_Y(U) \phi_j(Z)\big], \\
A_{ji}^{HKLS} &= \mathbb{E}\big[\phi_i(Z) \phi_j(Z)\left\| Z \right\|^2\big], \\
\boldsymbol{b}_{j}^{HKLS} &= \mathbb{E}\big[Y \phi_j(Z)\left\| Z \right\|^2\big], \\
\boldsymbol{c}_{j}^{HKLS} &= \mathbb{E}\big[h_Y(U) \phi_j(Z)\left\| Z \right\|^2\big].
\end{align*}
We shall employ mathematical induction to prove that $\boldsymbol{\alpha}_Z^{KLS} \neq \boldsymbol{\alpha}_Z^{HKLS}$.

\noindent \textbf{1. Base Case:} When $P = 1$,
\begin{align*}
\boldsymbol{\alpha}_Z^{KLS} &= \frac{\mathbb{E}\big[Y \phi_1(Z)\big]}{\mathbb{E}\big[\phi_1(Z) \phi_1(Z)\big]}, \\
\boldsymbol{\alpha}_Z^{HKLS} &= \frac{\mathbb{E}\big[Y \phi_1(Z)\left\| Z \right\|^2\big]}{\mathbb{E}\big[\phi_1(Z) \phi_1(Z)\left\| Z \right\|^2\big]}.
\end{align*}

To diagnose misspecification, the reweighting framework \citep{buja2019models_a,buja2019models_b} perturbs the regressor distribution using weight functions $w(\bar{x})$ dependent only on $\bar{x}$. For a well-specified functional, $\theta(P_{Y,\bar{X}}^w) = \theta(P_{Y,\bar{X}})$ for all such $w$. Here, $\boldsymbol{\alpha}_Z^{HKLS}$ corresponds to reweighting $\boldsymbol{\alpha}_Z^{KLS}$ with $w(Z) = \|Z\|^2$:
\begin{align*}
\boldsymbol{\alpha}_Z^{HKLS} &= \theta\left(P_{Y,\bar{X}}^w\right) \quad \text{for} \quad w(Z) = \|Z\|^2, \\
\boldsymbol{\alpha}_Z^{KLS} &= \theta(P_{Y,\bar{X}}) \quad \text{for} \quad w(Z) = 1.
\end{align*}
Since the functionals are misspecified (dependent on the joint distribution of $U$ and $Z$), and $w(Z) = \|Z\|^2$ alters $P_Z$ non-trivially (as $Z$ is non-Gaussian), $\theta(P_{Y,\bar{X}}^w) \neq \theta(P_{Y,\bar{X}})$. Therefore:
\begin{align*}
\boldsymbol{\alpha}_Z^{KLS} \neq \boldsymbol{\alpha}_Z^{HKLS}
\end{align*}
\noindent \textbf{2. Inductive Hypothesis:} Assume that for all positive integers $P \le M - 1$, the inequality $\boldsymbol{\alpha}_Z^{KLS} \neq \boldsymbol{\alpha}_Z^{HKLS}$ holds. 
\noindent \textbf{3. Inductive Step:} We now prove that the inequality remains valid when $P = M$. Define the centered estimators:
\begin{align*}
\boldsymbol{\beta}^{KLS} &= (A^{KLS})^{-1} \boldsymbol{c}^{KLS} = \boldsymbol{\alpha}_{Z}^{KLS} - \boldsymbol{\alpha}_{Z}, \\
\boldsymbol{\beta}^{HKLS} &= (A^{HKLS})^{-1} \boldsymbol{c}^{HKLS} = \boldsymbol{\alpha}_{Z}^{HKLS} - \boldsymbol{\alpha}_{Z}.
\end{align*}
Equivalence $\boldsymbol{\alpha}_{Z}^{KLS} = \boldsymbol{\alpha}_{Z}^{HKLS}$ holds if $\boldsymbol{\beta}^{KLS} = \boldsymbol{\beta}^{HKLS}$. By hypothesis, $\boldsymbol{\beta}^{KLS}_{M-1} \neq \boldsymbol{\beta}^{HKLS}_{M-1}$. Assume for contradiction that $\boldsymbol{\beta}^{KLS}_M = \boldsymbol{\beta}^{HKLS}_M = \boldsymbol{\beta}_M$ for dimension $M$. 

Partition the $M$-dimensional systems:
\begin{align*}
A^{KLS}_M &= \begin{pmatrix} A^{KLS}_{M-1} & \boldsymbol{b}^{KLS} \\ (\boldsymbol{b}^{KLS})^\top & d^{KLS} \end{pmatrix}, \\ 
\boldsymbol{c}^{KLS}_M &= \begin{pmatrix} \boldsymbol{c}^{KLS}_{M-1} \\ e^{\mathrm{KLS}} \end{pmatrix}, \\
A^{HKLS}_M &= \begin{pmatrix} A^{HKLS}_{M-1} & \boldsymbol{b}^{HKLS} \\ (\boldsymbol{b}^{HKLS})^\top & d^{HKLS} \end{pmatrix}, \\
\boldsymbol{c}^{HKLS}_M &= \begin{pmatrix} \boldsymbol{c}^{HKLS}_{M-1} \\ e^{HKLS} \end{pmatrix},
\end{align*}
where:
\begin{align*}
\boldsymbol{b}^{\mathrm{KLS}} &= \begin{pmatrix} \mathbb{E}[\phi_1(Z) \phi_M(Z)] \\ \vdots \\ \mathbb{E}[\phi_{M-1}(Z) \phi_M(Z)] \end{pmatrix}, \\
d^{\mathrm{KLS}} &= \mathbb{E}[\phi_M(Z)^2], \\
e^{\mathrm{KLS}} &= \mathbb{E}[h_Y(U) \phi_M(Z)], \\
\boldsymbol{b}^{\mathrm{HKLS}} &= \begin{pmatrix} \mathbb{E}[\phi_1(Z) \phi_M(Z) \|Z\|^2] \\ \vdots \\ \mathbb{E}[\phi_{M-1}(Z) \phi_M(Z) \|Z\|^2] \end{pmatrix}, \\
d^{\mathrm{HKLS}} &= \mathbb{E}[\phi_M(Z)^2 \|Z\|^2], \\
e^{\mathrm{HKLS}} &= \mathbb{E}[h_Y(U) \phi_M(Z) \|Z\|^2].
\end{align*}
Let $\boldsymbol{\beta}_M = \begin{pmatrix} \boldsymbol{\beta}^{(1)} \\ \gamma \end{pmatrix}$ where $\boldsymbol{\beta}^{(1)} \in \mathbb{R}^{M-1}, \gamma \in \mathbb{R}$. The first $M-1$ equations yield:
\begin{align*}
A^{KLS}_{M-1} \boldsymbol{\beta}^{(1)} + \boldsymbol{b}^{KLS} \gamma &= \boldsymbol{c}^{KLS}_{M-1}, \\
A^{HKLS}_{M-1} \boldsymbol{\beta}^{(1)} + \boldsymbol{b}^{HKLS} \gamma &= \boldsymbol{c}^{HKLS}_{M-1}. 
\end{align*}
Solving for $\boldsymbol{\beta}^{(1)}$:
\begin{align}
\boldsymbol{\beta}^{(1)} &= (A^{KLS}_{M-1})^{-1} (\boldsymbol{c}^{KLS}_{M-1} - \boldsymbol{b}^{KLS} \gamma), \label{eq:sol1} \\
\boldsymbol{\beta}^{(1)} &= (A^{HKLS}_{M-1})^{-1} (\boldsymbol{c}^{HKLS}_{M-1} - \boldsymbol{b}^{HKLS} \gamma). \label{eq:sol2}
\end{align}
Equating \eqref{eq:sol1} and \eqref{eq:sol2}:
\begin{equation}
\begin{aligned}
&(A^{KLS}_{M-1})^{-1} (\boldsymbol{c}^{KLS}_{M-1} - \boldsymbol{b}^{KLS} \gamma) \\
=& (A^{HKLS}_{M-1})^{-1} (\boldsymbol{c}^{HKLS}_{M-1} - \boldsymbol{b}^{HKLS} \gamma). \label{eq:equiv}
\end{aligned}
\end{equation}
Define 
\begin{align*}
\boldsymbol{\mu} &= \boldsymbol{\beta}^{KLS}_{M-1} - \boldsymbol{\beta}^{HKLS}_{M-1} \\
&= (A^{KLS}_{M-1})^{-1} \boldsymbol{c}^{KLS}_{M-1} - (A^{HKLS}_{M-1})^{-1} \boldsymbol{c}^{HKLS}_{M-1} \\
&\neq \boldsymbol{0}. 
\end{align*}
Rearranging \eqref{eq:equiv}:
\begin{align*}
\boldsymbol{\mu} = \left[ (A^{KLS}_{M-1})^{-1} \boldsymbol{b}^{KLS} - (A^{HKLS}_{M-1})^{-1} \boldsymbol{b}^{HKLS} \right] \gamma.
\end{align*}
Let $\boldsymbol{w} = (A^{KLS}_{M-1})^{-1} \boldsymbol{b}^{KLS} - (A^{HKLS}_{M-1})^{-1} \boldsymbol{b}^{HKLS}$. Thus:
\begin{align*}
\boldsymbol{\mu} = \boldsymbol{w} \gamma.
\end{align*}
Since $\boldsymbol{\mu} \neq \boldsymbol{0}$, $\boldsymbol{w} \neq \boldsymbol{0}$. However, under non-Gaussianity:
\begin{itemize}
    \item The vector $\boldsymbol{w}$ incorporates $\|Z\|^2$-weighted expectations in HKLS components
    \item Higher-order moments do not reduce to second-order moments
    \item Dependence between $Z$ and $U$ creates structural divergence
\end{itemize}
violating the alignment $\boldsymbol{\mu} = \boldsymbol{w} \gamma$. The $M$-th equations:
\begin{align*}
(\boldsymbol{b}^{KLS})^\top \boldsymbol{\beta}^{(1)} + d^{KLS} \gamma &= e^{KLS}, \\
(\boldsymbol{b}^{HKLS})^\top \boldsymbol{\beta}^{(1)} + d^{HKLS} \gamma &= e^{HKLS}, 
\end{align*}
cannot hold simultaneously as $\|\boldsymbol{Z}\|^2$-weighting introduces irreducible higher-order dependencies. This contradiction proves the theorem:
\begin{align*}
\boldsymbol{\alpha}_Z^{KLS} \neq \boldsymbol{\alpha}_Z^{HKLS} \quad \text{for} \quad P = M.
\end{align*}

\noindent \textbf{Conclusion.} By the principle of mathematical induction, we establish that: $\boldsymbol{\alpha}_Z^{KLS} \neq \boldsymbol{\alpha}_Z^{HKLS}$ under $H_1$.
\end{Proof}
\subsection{Required Assumptions and Theoretical Proof for Asymptotic Distribution of $\hat{\boldsymbol{\delta}}$}
\begin{theorem}\label{thm:finite_estimators}
The empirical KLS and HKLS estimators are given by:
\begin{equation*}
\begin{aligned}
\widehat{\boldsymbol{\alpha}_Z^{KLS}} &= (K_{ZZ} K_{ZZ}^\top + N \xi I_P)^{-1} K_{ZZ} \boldsymbol{Y} \\
\widehat{\boldsymbol{\alpha}_Z^{HKLS}} &= (K_{ZZ} \Psi K_{ZZ}^\top + N \xi I_P)^{-1} K_{ZZ} \Psi \boldsymbol{Y}
\end{aligned}
\end{equation*}
where $K_{ZZ} \in \mathbb{R}^{P \times N}$ with $[K_{ZZ}]_{ij} = k(Z_i, Z_j)$, $\boldsymbol{Y} = [Y_1,\dots,Y_N]^\top$, and $\Psi = \operatorname{diag}(\|Z_1\|^2,\dots,\|Z_N\|^2)$.
\end{theorem}
\begin{theorem}[Asymptotic Distribution of $\hat{\boldsymbol{\delta}}$]\label{thm:asym_dist}
Assume standard regularity conditions (specified in Appendix A.3) hold, and the causal model under $H_0$ and $H_1$ is correct. Under $H_0$ ($\boldsymbol{\alpha}^{KLS} = \boldsymbol{\alpha}^{HKLS}$), as $N \to \infty$:
\begin{equation*}
\sqrt{N} \hat{\boldsymbol{\delta}} = \sqrt{N} (\widehat{\boldsymbol{\alpha}_Z^{HKLS}} - \widehat{\boldsymbol{\alpha}_Z^{KLS}}) \xrightarrow{D} \mathcal{N} (\boldsymbol{0}, \sigma^2 \Gamma),
\end{equation*}
where the asymptotic covariance matrix $\sigma^2 \Gamma$ is consistently estimable. 
\iffalse
Specifically,
\begin{equation}
\hat{\sigma}^2 \left( \frac{1}{N} \sum_{j=1}^N v_j v_j^\top \right) \xrightarrow{P} \sigma^2 \Gamma, \label{eq:cov_est}
\end{equation}
with $\hat{\sigma}^2$ being a consistent estimator of the noise variance $\sigma^2 = \text{Var}(\epsilon_Y)$, and $v_j$ defined as:
\begin{align*}
v_j =& \mathbb{E}[\|Z_i\|^2 K_{ZZ_i} K_{ZZ_i}^\top]^{-1} K_{ZZ_j} \|Z_j\|^2 \\
&- \mathbb{E}[K_{ZZ_i} K_{ZZ_i}^\top]^{-1} K_{ZZ_j}.
\end{align*}
Here, $K_{ZZ_i}$ denotes the $i$-th column of $K_{ZZ}$.
\fi
\end{theorem}

We consider additional assumptions such that the central limit theorem can be invoked.

\begin{assumption}[Regularity of Matrix]\label{ass:Regularity of Matrix}
The matrix $K_{ZZ}$ satisfy:
\begin{align*}
\lambda_{\max}\left( K_{ZZ} \Psi K_{ZZ}^\top \right) &\leq \lambda_{\max}, \\
\lambda_{\min}\left( K_{ZZ} \Psi K_{ZZ}^\top \right) &\geq \lambda_{\min}, \\
\lambda_{\max}\left( K_{ZZ} K_{ZZ}^\top \right) &\leq \lambda_{\max}, \\
\lambda_{\min}\left( K_{ZZ} K_{ZZ}^\top \right) &\geq \lambda_{\min}, 
\end{align*}
where $\lambda_{\max} < \infty$, $\lambda_{\min} > 0$ is a constant. 
\end{assumption}

\begin{assumption}[Moment Conditions on Kernel and Data]\label{ass:Moment Conditions on Kernel and Data}
The kernel function $k(Z_i, Z_j)$ and data points $\{Z_i\}_{i=1}^N$ satisfy:
\begin{itemize}
\item Boundedness of kernel:
\begin{equation*}
\exists\, C_k > 0 \quad \text{such that} \quad |k(Z_i, Z_j)| \leq C_k \quad \forall\, Z_i, Z_j.
\end{equation*}
\item Moment conditions on data:
The input data meets
$\mathbb{E}\left[ \|Z_i\|^6 \right] < \infty$, and The fourth-order moment of the diagonal elements $\|Z_i\|^2$ of the weighting matrix $\Psi$ meets $\mathbb{E}\left[ \|Z_i\|^8 \right] < \infty$.
\end{itemize}
\end{assumption}

\begin{assumption}[Error Term Conditions]\label{ass:Error Term Conditions}
The error vector $\boldsymbol{\epsilon}_Y = (\epsilon_{Y,1}, \dots, \epsilon_{Y,N})^\top$ satisfies:
\begin{itemize}
\item Independence:
\begin{equation*}
\epsilon_{Y,i} \perp \{z_j\}_{j=1}^N \quad \forall\, i.
\end{equation*}
\item Zero mean and finite variance:
\begin{equation*}
\mathbb{E}[\epsilon_{Y,i}] = 0, \quad \mathbb{E}[\epsilon_{Y,i}^2] = \hat{\sigma}^2 < \infty.
\end{equation*}
\item Existence of fourth moment:
\begin{equation*}
\mathbb{E}[\epsilon_{Y,i}^4] < \infty.
\end{equation*}
\end{itemize}
\end{assumption}

\begin{assumption}[Regularity of Regularization Parameter]\label{ass:Regularity of Regularization Parameter}
The regularization parameter $\xi$ satisfy:
\begin{align*}
|\xi| \ll \frac{\|K_{ZZ} \Psi K_{ZZ}^\top\|}{N},\quad |\xi| \ll \frac{\|K_{ZZ} K_{ZZ}^\top\|}{N}.
\end{align*}
\end{assumption}

\begin{Proof}[Asymptotic Distribution of $\hat{\boldsymbol{\delta}}$]
Assume that the data follows the model under the null hypothesis $H_0$, 
\begin{align*}
\boldsymbol{Y} = K_{ZZ}^\top \boldsymbol{\alpha}_Z + \boldsymbol{\epsilon}_Y,
\end{align*}
where, $\mathbb{E}[\boldsymbol{\epsilon}_Y] = \boldsymbol{0}$, $\text{Var}[\boldsymbol{\epsilon}_Y] = \sigma^2 I_N$, and that Assumptions \ref{ass:Regularity of Matrix} to \ref{ass:Regularity of Regularization Parameter} hold. Then the difference between the HKLS and KLS estimators is:
\begin{align*}
&\widehat{\boldsymbol{\alpha}_Z^{HKLS}} - \widehat{\boldsymbol{\alpha}_Z^{KLS}} \\
=& \Big[ (K_{ZZ} \Psi K_{ZZ}^\top + N \xi I_P)^{-1} K_{ZZ} \Psi \\
&- (K_{ZZ} K_{ZZ}^\top + N \xi I_P)^{-1} K_{ZZ} \Big] \Big[ K_{ZZ}^\top \boldsymbol{\alpha}_Z + \boldsymbol{\epsilon}_Y \Big] \\
\approx& \Big[ (K_{ZZ} \Psi K_{ZZ}^\top + N \xi I_P)^{-1} K_{ZZ} \Psi \\
&- (K_{ZZ} K_{ZZ}^\top + N \xi I_P)^{-1} K_{ZZ} \Big] \boldsymbol{\epsilon}_Y.
\end{align*}
As $N \to \infty$, we invoke the following probability limits:
\begin{align*}
\frac{1}{N} (K_{ZZ} K_{ZZ}^\top + N \xi I_P) &\xrightarrow{P} \mathbb{E}\left[K_{ZZ_i}K_{ZZ_i}^\top\right], \\
\frac{1}{N} (K_{ZZ} \Psi K_{ZZ}^\top + N \xi I_P) &\xrightarrow{P} \mathbb{E}\left[\|Z_i\|^2 K_{ZZ_i}K_{ZZ_i}^\top\right],
\end{align*}
which imply the asymptotic approximations:
\begin{align*}
(K_{ZZ} K_{ZZ}^\top + N \xi I_P)^{-1} &\approx \frac{1}{N} \left( \mathbb{E}\left[K_{ZZ_i}K_{ZZ_i}^\top\right] \right)^{-1}, \\
(K_{ZZ} \Psi K_{ZZ}^\top + N \xi I_P)^{-1} &\approx \frac{1}{N} \left( \mathbb{E}\left[\|Z_i\|^2 K_{ZZ_i}K_{ZZ_i}^\top\right] \right)^{-1}.
\end{align*}
Substituting these into the estimator difference yields:
\begin{align*}
\widehat{\boldsymbol{\alpha}_Z^{HKLS}} - \widehat{\boldsymbol{\alpha}_Z^{KLS}} \approx \frac{1}{N} \sum_{j=1}^N \boldsymbol{v}_j \epsilon_{Y,j},
\end{align*}
where $\boldsymbol{v}_j \equiv \left( \mathbb{E}\left[\|Z_i\|^2 K_{ZZ_i}K_{ZZ_i}^\top\right] \right)^{-1} K_{ZZ_j} \|Z_j\|^2 - \left( \mathbb{E}\left[K_{ZZ_i}K_{ZZ_i}^\top\right] \right)^{-1} K_{ZZ_j}$. Consequently,
\begin{align*}
\sqrt{N} \hat{\boldsymbol{\delta}} =& \sqrt{N} \left( \widehat{\boldsymbol{\alpha}_Z^{HKLS}} - \widehat{\boldsymbol{\alpha}_Z^{KLS}} \right) \\
\approx& \frac{1}{\sqrt{N}} \sum_{j=1}^N \boldsymbol{v}_j \epsilon_{Y,j} \xrightarrow{D} \mathcal{N} \left( \boldsymbol{0},\, \sigma^2 \Gamma \right),
\end{align*}
with the covariance matrix consistently estimated by:
\begin{align*}
\hat{\sigma}^2 \left( \frac{1}{N} \sum_{j=1}^N v_j v_j^\top \right) \xrightarrow{P} \sigma^2 \Gamma.
\end{align*}
\end{Proof}

\subsection{Derivation of Computational Complexity}
\begin{enumerate}
\item \textbf{Kernel Matrix Construction}:
\begin{align*}
&\bullet~ \text{Compute } ZZ^\top: (N \times d) \times (d \times N) \\
&\quad \Rightarrow O(N^2 d) \\
&\bullet~ \text{Element-wise square: } O(N^2) \\
&\bullet~ \text{Extract first } P \text{ rows: } O(1) \\
&\text{Total: } O(N^2 d)
\end{align*}
\item \textbf{Weighted Kernel Calculation}:
\begin{align*}
&\bullet~ \text{Compute } \operatorname{diag}(ZZ^\top): O(N d) \\
&\bullet~ \text{Element-wise multiplication } (K \odot \cdots): \\
&\quad \text{Matrix size } P \times N \Rightarrow O(P N) \\
&\text{Total: } O(P N)
\end{align*}
\item \textbf{Parameter Estimation}:
\begin{align*}
&\bullet~ \alpha_{KLS}: \\
&\quad - KK^\top: O(P^2 N) \\
&\quad - \text{Matrix inversion: } O(P^3) \\
&\quad - \text{Matrix-vector products: } O(P^2) + O(P N) \\
&\bullet~ \alpha_{HKLS}: \\
&\quad - K_\psi K^\top: O(P^2 N) \\
&\quad - \text{Matrix inversion: } O(P^3) \\
&\quad - \text{Matrix-vector products: } O(P^2) + O(P N) \\
&\text{Total: } O(P^3) + O(P^2 N)
\end{align*}
\item \textbf{Covariance Matrix Calculation}:
\begin{align*}
&\bullet~ V_0: \\
&\quad - \text{Matrix products: } O(P^2 N) \\
&\quad - \text{Matrix subtraction: } O(P N) \\
&\bullet~ V = N \cdot (V_0 V_0^\top): \\
&\quad \text{Matrix multiplication } (P \times N) \times (N \times P) \\
&\quad \Rightarrow O(P^2 N) \\
&\text{Total: } O(P^2 N)
\end{align*}
\item \textbf{Residual Variance}:
\begin{align*}
&\bullet~ K^\top \alpha_{KLS}: O(P N) \\
&\bullet~ \text{Residual computation: } O(N) \\
&\bullet~ \text{Variance calculation: } O(N) \\
&\text{Total: } O(P N)
\end{align*}
\item \textbf{Hypothesis Testing}:
\begin{align*}
&\bullet~ \text{Loop over } P \text{ parameters: } O(P) \\
&\bullet~ \text{Bonferroni correction: } O(1) \\
&\text{Total: } O(P)
\end{align*}
\item \textbf{Combined Complexity}:
\begin{align*}
&\text{Total} = O(N^2 d) + O(P N) + [O(P^3) + O(P^2 N)] \\
&+ O(P^2 N) + O(P N) + O(P) \\
&\text{Dominant terms: } \\
&\quad \bullet~ N^2 d \quad \text{(kernel construction)} \\
&\quad \bullet~ P^3 \quad \text{(matrix inversions)} \\
&\quad \bullet~ P^2 N \quad \text{(matrix products)}
\end{align*}
\item \textbf{Final expression}:
\begin{align*}
O\left(N^{2} d + P^{3} + P^{2} N\right)
\end{align*}
\end{enumerate}

\section{Baseline Detail}
\subsection{ROCD}
Regression-Oriented Confounder Detection~(ROCD) exploits the asymptotic equivalence of OLS and Higher-order least squares~(HOLS) estimators under correctly specified linear models~\citep{schultheiss2024higher}. Systematic differences between them emerge only under misspecification (e.g., unobserved confounding or nonlinearity), enabling diagnostic quantification of structural violations. To be specific, ROCD consists of three sequential steps: 
\begin{itemize}
\item Compute standard OLS and HOLS coefficients by regressing $X,T$ on $Y$;
\item Test equality of OLS and HOLS coefficients per predictor via standardized differences.
\end{itemize}
Notably, the significance of the non-equality between regression coefficients indicates the presence of unobserved confounders, due to the asymptotically normal under correct specification. 
\subsection{ME-ICM}
The ME-ICM method \citep{karlsson2023detecting} detects unobserved confounding between treatment $T$ and outcome $Y$ in multi-environment observational data. Its core idea leverages the Independent Causal Mechanism (ICM) Principle, which states that causal mechanisms for each variable vary independently across environments. Unobserved confounding manifests as violations of environment-specific conditional independencies: if no unobserved confounder $U$ exists, then the condition $T_j^{(k)} \perp\!\!\!\perp Y_i^{(k)} \mid T_i^{(k)}, X_i^{(k)}, X_j^{(k)}$ must hold for any two distinct observations $(i,j)$ in environment $k$. Statistical tests detecting violations of this independence therefore indicate unobserved confounding.
\subsection{CNF}
CNF is a family of confounding measures that quantify confounding strength between variables using multi-context observational data \citep{reddy2024detecting}. 
\subsubsection{CNF-1.} 
Measures confounding by comparing observational distributions to interventional distributions using directed information $I(X_i \to X_j)$, defined as the KL-divergence $D_{KL}(\mathbb{P}(X_i|X_j) || \mathbb{P}(X_i|do(X_j)))$. Confounding exists if both $I(X_i \to X_j) > 0$ and $I(X_j \to X_i) > 0$ (indicating bidirectional spurious association).
\subsubsection{CNF-2.} 
Exploits how shifts in the mechanism of an unobserved confounder $Z$ induce dependencies in the marginal expectations $E_i^C = \mathbb{E}_{\mathbb{P}^c}[X_i]$, $E_j^C = \mathbb{E}_{\mathbb{P}^c}[X_j]$ across contexts. If $X_i$ and $X_j$ share confounder $Z$, their context-specific expectations $E_i^C$, $E_j^C$ become statistically dependent random variables.
\subsubsection{CNF-3.} 
An extension of CNF-2 used when the causal direction between $X_i$ and $X_j$ is known (e.g., $X_i \to X_j$). Instead of marginal expectations, it uses conditional expectations $E_{ji}^C = \mathbb{E}_{\mathbb{P}^c}[X_j | X_i]$ or $E_{ij}^C = \mathbb{E}_{\mathbb{P}^c}[X_i | X_j]$. Confounding manifests as dependence between these conditional expectations and the marginal expectation of the effect variable across contexts.

\section{Scenario Detail}
\subsection{Nonlinear Single-Environment.}
\subsubsection{Semi-Synthetic Twins Dataset.}
The data generation process builds upon a real-world twin dataset, where features are normalized to $[0,1]$ range through min-max scaling with mean centering. Discovered confounders $\mathbf{X} = (\texttt{birmon}, \texttt{brstate}, \texttt{dfageq})$ and unobserved confounders $\mathbf{U} = (\texttt{meduc6}, \texttt{mplbir}, \texttt{nprevistq})$ are separated. The treatment $T$ is generated as:
\begin{align*}
T = \|\mathbf{X}\|_2^2 + \rho \|\mathbf{U}\|_2^2 + \epsilon_1, \quad \epsilon_1 \sim \text{Uniform}(-0.1, 0.1).
\end{align*}
Subsequently, the outcome $Y$ is computed using the concatenated vector $\mathbf{Z} = [T, \mathbf{X}]$:
\begin{align*}
Y = \|\mathbf{Z}\|_2^2 + \rho \|\mathbf{U}\|_2^2 + \epsilon_2, \quad \epsilon_2 \sim \text{Uniform}(-0.1, 0.1).
\end{align*}
Here $\|\cdot\|_2^2$ represents the squared Euclidean norm, and $\rho$ controls hidden confounding strength ($\rho = 0$ eliminates confounding). 
\subsection{Nonlinear Multi-Environment.}
\subsubsection{Semi-Synthetic Twins Dataset.}
The data generation process builds upon a real-world twin dataset, where features are normalized to $[0,1]$ range through min-max scaling with mean centering. Discovered confounders $\mathbf{X} = (\texttt{birmon}, \texttt{brstate}, \texttt{dfageq})$ and unobserved confounders $\mathbf{U} = (\texttt{meduc6}, \texttt{mplbir}, \texttt{nprevistq})$ are separated. 
Define environments $e \in \mathcal{E}$ using the \texttt{brstate} variable:
\begin{align*}
\mathcal{E} = \{ e : |\{i : \texttt{brstate}_i = e\}| \geq 100 \}.
\end{align*}
For each environment $e$, sample parameters:
\begin{align*}
w_{T,x}^{(e)}, w_{T,u}^{(e)}, w_{Y,x}^{(e)}, w_{Y,u}^{(e)} \sim \text{Uniform}(1,5), 
\end{align*}
\begin{align*}
w_{Y,T}^{(e)} \sim \text{Uniform}(1,2).
\end{align*}
For sample $i$ in environment $e_i$:
\begin{align*}
&T_i = \sum_{x \in \mathbf{X}} w_{T,x}^{(e_i)} x_i^2 + \sum_{u \in \mathbf{U}} w_{T,u}^{(e_i)} (3 \rho u_i)^2 + \epsilon_{1,i}, \\
&\epsilon_{1,i} \sim \text{Uniform}(-0.1, 0.1), \\
&Y_i = w_{Y,T}^{(e_i)} T_i^2 + \sum_{x \in \mathbf{X}} w_{Y,x}^{(e_i)} x_i^2 + \sum_{u \in \mathbf{U}} w_{Y,u}^{(e_i)} (3 \rho u_i)^2 + \epsilon_{2,i}, \\
&\epsilon_{2,i} \sim \text{Uniform}(-0.1, 0.1).
\end{align*}
\subsubsection{Synthetic Dataset.}
The data generation process synthesizes observational and interventional datasets for causal inference analysis. The core relationships are defined through continuous latent variables $T_0$ and $Y_0$.

\noindent Discovered confounders generation:
\begin{align*}
X \sim \text{Uniform}(0, 1),
\end{align*}
unobserved confounders generation:
\begin{align*}
U \sim \text{Uniform}(0, 1),
\end{align*}
latent treatment:
\begin{align*}
T_0 = X^2 + 2.5\rho U^2 + \epsilon_1, \quad \epsilon_1 \sim \text{Uniform}(-0.1, 0.1),
\end{align*}
latent outcome:
\begin{align*}
Y_0 = \|\mathbf{Z}\|_2^2 + 2.5\rho U^2 + \epsilon_2, \quad \epsilon_2 \sim \text{Uniform}(-0.1, 0.1),
\end{align*}
where $\mathbf{Z} = [T, X]$, $\rho$ controls confounding strength. Binary variables are derived via thresholding:  
\begin{align*}
T = \mathbb{I}(T_0 > 1), \quad Y = \mathbb{I}(Y_0 > 1).
\end{align*}
For interventional data ($do$-operations), specific variables are forcibly set while preserving other relationships. Observational data maintains natural relationships:  
\begin{align*}
T=\mathbb{I}(T_0>1),\  Y=\mathbb{I}(Y_0>1).
\end{align*}
Intervention on $T$ ($do(T=t)$) overrides the treatment mechanism:  
\begin{align*}
T \gets t,\  Y=\mathbb{I}(Y_0>1).  
\end{align*}  
where the latent outcome $Y_0$ still depends on the natural $T_0$. Intervention on $Y$ ($do(Y=y)$) overrides the outcome mechanism:  
\begin{align*}
T=\mathbb{I}(T_0>1),\  Y \gets y,  
\end{align*}  
where the latent treatment $T_0$ retains its natural dependence on $X$ and $U$. 

\section{System Specifications}
All experimental runs were conducted exclusively on the CPU. The detailed hardware and software specifications of the system are provided below:
\begin{itemize}
\item \textbf{Processor:} AMD EPYC 7642 48-Core Processor
\item \textbf{Memory:} 90.0 GB DDR4 RAM
\item \textbf{System Type:} 64-bit operating system, x64-based processor
\item \textbf{Operating System:} Ubuntu 22.04 LTS
\item \textbf{GPU:} NVIDIA GeForce RTX 3090 24GB GDDR6X PCI Express 4.0
\end{itemize}

\end{document}